\documentclass[10pt,twocolumn,letterpaper]{article}

\usepackage[pagenumbers]{cvpr} 

\usepackage[dvipsnames]{xcolor}

\usepackage{colortbl}
\usepackage{amsthm}
\usepackage{multirow}
\usepackage{algorithm}
\usepackage{algorithmic}

\newcommand{\Frechet}{Fr\'echet}
\newcommand{\tworow}[2]{\begin{tabular}{@{}l@{}}#1 \\ #2\end{tabular}}

\usepackage{pifont}
\newcommand{\cmark}{\ding{51}}%
\newcommand{\xmark}{\ding{55}}%

\newcommand{\R}{\mathbb{R}}

\newcommand{\Ds}{D_s}
\newcommand{\Dt}{D_t}
\newcommand{\Dst}{D_{s\rightarrow t}}

\theoremstyle{definition}

\definecolor{cvprblue}{rgb}{0.21,0.49,0.74}
\usepackage[pagebackref,breaklinks,colorlinks,allcolors=cvprblue]{hyperref}

\def\paperID{*****} 
\def\confName{CVPR}
\def\confYear{2025}

\title{\Frechet{} Radiomic Distance (FRD): A Versatile Metric for Comparing Medical Imaging Datasets}

\author{
Nicholas Konz$^{1}$\thanks{Equal contribution.} \\
{\tt\small nicholas.konz@duke.edu}
\and
Richard Osuala$^{2,3,4}$\footnotemark[1] \\
{\tt\small richard.osuala@ub.edu}
\and
Preeti Verma$^{2}$ \\
\and
Yuwen Chen$^{1}$ \\
\and
Hanxue Gu$^{1}$ \\
\and
Haoyu Dong$^{1}$ \\
\and
Yaqian Chen$^{1}$ \\
\and
Andrew Marshall$^{8}$ \\
\and
Lidia Garrucho$^{2}$ \\
\and
Kaisar Kushibar$^{2}$ \\
\and
Daniel M. Lang$^{3,4}$ \\
\and
Gene S. Kim$^{9}$ \\
\and
Lars J. Grimm$^{5}$ \\
\and
John M. Lewin$^{10}$ \\
\and
James S. Duncan$^{8,10,11}$ \\
\and
Julia A. Schnabel$^{3,4,12}$ \\
\and
Oliver Diaz$^{2,13}$ \\
\and
Karim Lekadir$^{2,14}$ \\
\and
Maciej A. Mazurowski$^{1,5,6,7}$ \\
\and
\textit{Full affiliations under \hyperref[sec:authoraffil]{Author Information}.}\\
\small{
\textbf{Code:} \url{https://github.com/RichardObi/frd-score}}\\
\small{
\textbf{Evaluation Framework:} \url{https://github.com/mazurowski-lab/medical-image-similarity-metrics}
}
}

\begin{document}
\maketitle

\begin{abstract}
Determining whether two sets of images belong to the same or different distributions or domains is a crucial task in modern medical image analysis and deep learning; for example, to evaluate the output quality of image generative models. Currently, metrics used for this task either rely on the (potentially biased) choice of some downstream task, such as segmentation, or adopt task-independent perceptual metrics (\eg, \Frechet{} Inception Distance/FID) from natural imaging, which we show insufficiently capture anatomical features. To this end, we introduce a new perceptual metric tailored for medical images, FRD (\Frechet{} Radiomic Distance), which utilizes standardized, clinically meaningful, and interpretable image features. We show that FRD is superior to other image distribution metrics for a range of medical imaging applications, including out-of-domain (OOD) detection, the evaluation of image-to-image translation (by correlating more with downstream task performance as well as anatomical consistency and realism), and the evaluation of unconditional image generation.
Moreover, FRD offers additional benefits such as stability and computational efficiency at low sample sizes, sensitivity to image corruptions and adversarial attacks, feature interpretability, and correlation with radiologist-perceived image quality. Additionally, we address key gaps in the literature by presenting an extensive framework for the multifaceted evaluation of image similarity metrics in medical imaging---including the first large-scale comparative study of generative models for medical image translation---and release an accessible codebase to facilitate future research. Our results are supported by thorough experiments spanning a variety of datasets, modalities, and downstream tasks, highlighting the broad potential of FRD for medical image analysis.
\end{abstract}

\section{Introduction}
\label{sec:intro}

Comparing image distributions is crucial in deep learning-driven medical image analysis. Example applications include out-of-domain (OOD) detection \citep{tschuchnig2022anomaly}, \eg, for detecting if new medical images were acquired using different protocols; the evaluation of image-to-image translation models, \eg, for converting MRI (magnetic resonance imaging) to CT (computed tomography) \citep{wolterink2017deep, armanious2020medgan}; the quality assessment of images generated to supplement real training data \citep{chen2021synthetic, pinaya2022brain}; and others \citep{chan2020deep}.

However, image distribution metrics from general computer vision (\eg, \Frechet{} Inception Distance/FID \citep{fid}) often miss key requirements for medical image analysis, and questioning of their unadapted use for this subfield has recently begun \citep{osuala2023data,woodland2024_fid_med,konz2024rethinking,wu2025pragmatic}. For example, in medical image OOD detection and image-to-image translation, the focus extends beyond just general image quality to \textit{image-level domain adaptation}: ensuring that source-domain images (\eg, from one scanner, vendor or institution) are compatible with diagnostic models trained on target-domain images from another scanner/vendor/institution, addressing ubiquitous domain shift issues 
in medical imaging \citep{durrer2024diffusion,beizaee2023harmonizing,wang2024mutual,modanwal2020mri,yang2019unsupervised,liu2021style,zhang2018translating,guan2021domain,yao2019baseline,maartensson2020reliability}. 
Additionally, medical imaging requires metrics that specifically capture anatomical consistency and realism, as well as clinical interpretability \citep{maier2024metrics,salahuddin2022transparency,chen2022explainable,singh2020explainable}. We argue that these specialized requirements are overlooked by the current metrics used for comparing sets of real and/or synthetic medical images.

\begin{figure}[htbp!]
    \centering
    \includegraphics[width=0.99\linewidth]{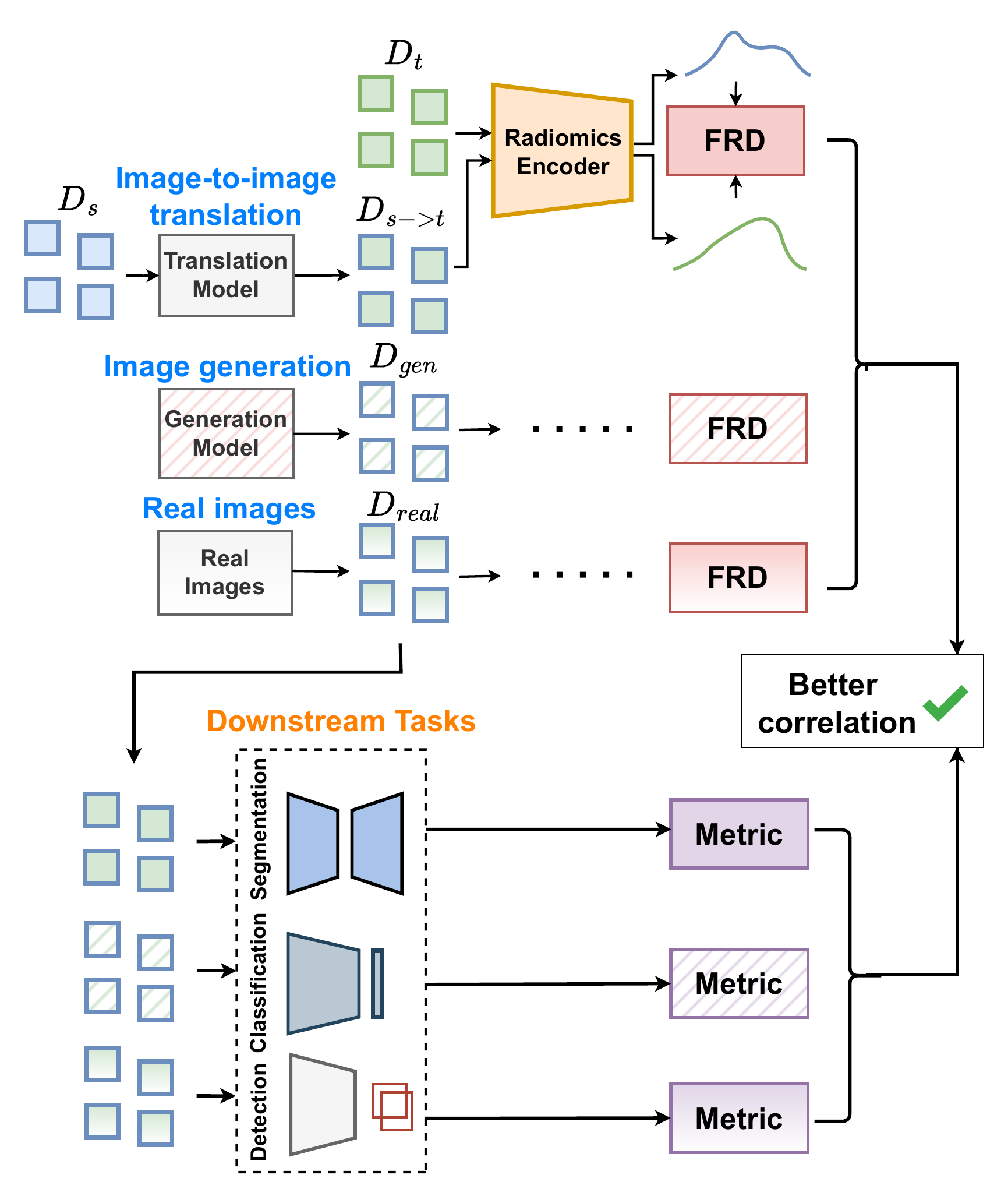}
    \caption{\textbf{Summary of our main contribution: }\textbf{FRD}, a metric designed from the ground up for comparing unpaired distributions of real and/or generated medical images.}
    \label{fig:teaser}
\end{figure}

The common approach of comparing medical image distributions in terms of the performance of some downstream task such as segmentation (because direct qualitative image assessment by radiologists is expensive and non-standardized) is driven by the choice of task, and requires costly training and labeling efforts. A task-independent metric that captures general image quality \textit{and} aligns with expected downstream task performance would therefore be preferable. In computer vision, perceptual metrics like FID are commonly used to evaluate image quality relative to real images \citep{fid,saxena2021comparison,binkowski2018demystifying}, yet these metrics are based on natural image features. Despite this, many applications of medical image translation \citep{li2023zero,wang2024mutual,li2023adaptive,shi2023mulhist} and generation \citep{pinaya2022brain,hashmi2024xreal,hansen2024inpainting,sun2022hierarchical} rely on FID (or the related Kernel Inception Distance/KID \citep{binkowski2018demystifying}) for evaluation, even though recent findings suggest these metrics may poorly reflect medical image quality \citep{segdiff,konz2024rethinking,contourdiff,wu2025pragmatic}. Our experiments further support this issue. Moreover, to date, no studies have proposed \textit{interpretable} metrics specifically tailored for comparing unpaired medical image distributions, despite the importance of explainability in medical image analysis.

In this paper, we showcase and address limitations in current metrics for comparing unpaired medical image distributions. We begin by evaluating ``RadiologyFID'' (RadFID) \citep{medigan}, a natural extension from FID which uses RadImageNet \citep{radimagenet} features instead of ImageNet \citep{deng2009imagenet} features, that has seen surprisingly little use. We find that RadFID improves upon prior metrics in some areas, yet lacks in interpretability, stability on small datasets, and other essential qualities. 

\begin{figure}[htbp!]
    \centering
    \includegraphics[width=0.95\linewidth]{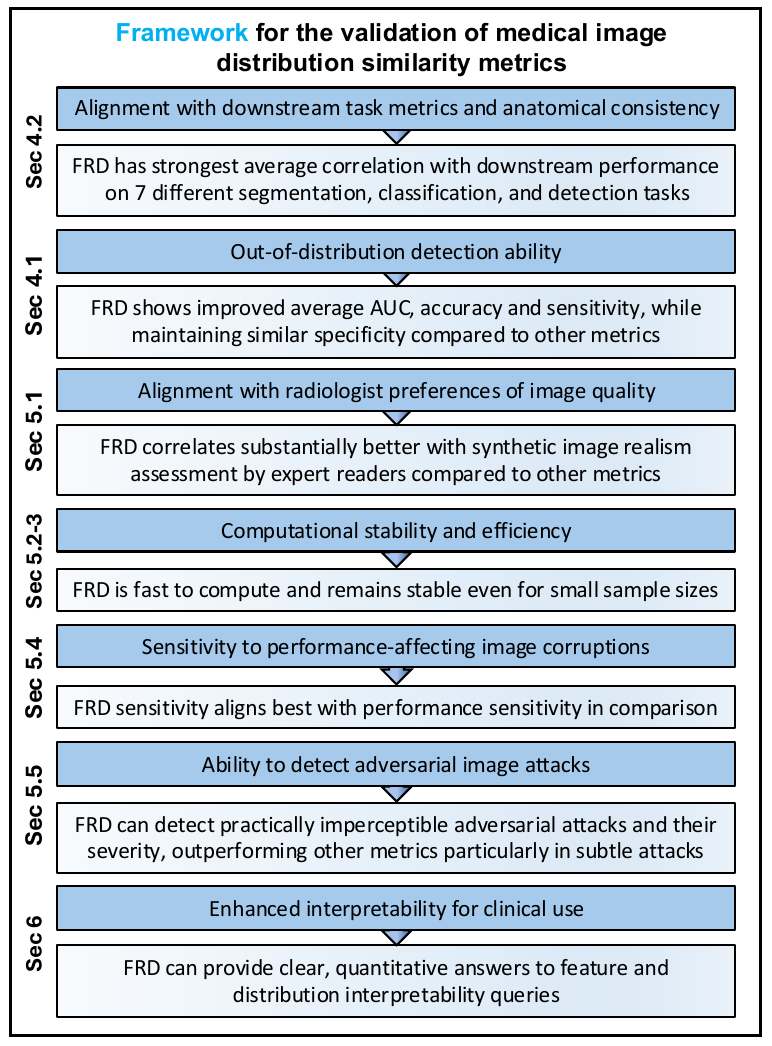}
    \caption{Our evaluation framework for FRD and other medical image distribution similarity metrics.}
    \label{fig:framework}
\end{figure}

To address these gaps, we introduce \Frechet{} Radiomic Distance (\textbf{FRD}), a metric leveraging pre-defined, interpretable \textit{radiomic} features which are widely used in medical image analysis (see \eg, \citet{yip2016applications,gillies2016radiomics,pyradiomics,lambin2012radiomics}). In addition to the inherent interpretability of the radiomic features it uses \citep{cui2023interpretable,orton2023interpretability,bang2021interpretable,ye2024radiomics,rifi2023interpretability}, FRD offers numerous advantages over learned feature metrics like FID and RadFID, which we demonstrate for various applications via an extensive evaluation framework for medical image distribution similarity metrics, summarized in Fig. \ref{fig:framework}. FRD is an improved version of our previous early version ``FRD$_\text{\textbf{v0}}$'' \citep{frd}, being both more robust and capturing a larger, more descriptive space of image features (Sec. \ref{sec:radiomicdist}); we further verify the resulting improvements in this work with far more extensive intrinsic and extrinsic evaluations than were present in our preliminary study.

We demonstrate these results in key application areas for unpaired medical image distribution comparison, such as out-of-domain (OOD) detection/analysis and the evaluation of image-to-image translation and image generation models. Our experiments span a wide range of medical image datasets and downstream tasks, image translation and generation models, and perceptual metrics. The datasets (see Table \ref{tab:datasets}) cover broad medical imaging scenarios that possess different image domains, including breast MRI inter-scanner data from different vendors, inter-sequence brain MRI data involving varying sequences, inter-modality data such as lumbar spine and abdominal MRI and CT, and others, all of which present unique challenges for the explored tasks. \textbf{We summarize our contributions as follows:}
\begin{enumerate}
    \item We highlight the shortcomings of common metrics for image distribution comparison (\eg, FID) in meeting the unique requirements of medical imaging.
    \item We introduce FRD, a task-independent perceptual metric based on radiomic features, which offers various improvements over prior metrics: \textbf{(1)} alignment with downstream tasks, \textbf{(2)} stability and computational efficiency for small datasets, \textbf{(3)} clinical interpretability, \textbf{(4)} sensitivity to image corruptions and adversarial attacks, and \textbf{(5)} alignment with radiologist perceptions of image quality.
    \item We validate FRD across diverse medical imaging datasets and applications, such as practical out-of-domain detection (including proposing a novel, standardized \textit{dataset-level} OOD metric), image-to-image translation,  image generation, and others, demonstrating its superiority in unpaired medical image distribution comparison.
    \item We show that, thanks to various improvements, including more robust feature normalization and the inclusion of many additional frequency space-based features which form a more expressive space of features to describe medical images, FRD outperforms our initially proposed FRD$_\text{\textbf{v0}}$ (as well as other metrics) in essentially all tested scenarios.
    \item Finally, we present a general-purpose evaluation framework for medical image similarity metrics, including the first large-scale comparative study of generative models for medical image-to-image translation, and release an accessible codebase to support future research. 
\end{enumerate}

We summarize FRD and visualize potential applications for it in Fig. \ref{fig:teaser}. FRD can be readily used and integrated into experiment pipelines via our accessible codebase packaged as a Python library at \url{https://github.com/RichardObi/frd-score}. In addition, our evaluation framework for medical image similarity metrics can be utilized at \url{https://github.com/mazurowski-lab/medical-image-similarity-metrics}.

\section{Related Work}
\label{sec:relwork}

\subsection{Metrics for Comparing Image Distributions} 
The standard approach for comparing two unpaired sets/distributions of images \( D_1, D_2 \subset \mathbb{R}^n \) involves defining a distance metric between them that satisfies basic properties (reflexivity, non-negativity, symmetry, and the triangle inequality) \citep{jayasumana2023rethinking}. In image-to-image translation for example, \( D_2 \) represents images translated from a source domain to a target domain, and \( D_1 \) is a set of real target domain images. For unconditional generation, \( D_2 \) contains generated images, and \( D_1 \) serves as a real reference set.

Typically, images from $D_1$ and $D_2$ are first encoded into a lower-dimensional feature space \( F_1, F_2 \subset \mathbb{R}^m \) respectively via an encoder \( f(x): \mathbb{R}^n \rightarrow \mathbb{R}^m \). Then, a distance such as the \Frechet{} distance \citep{frechet1957distance}---technically the 2-Wasserstein distance---is computed between these feature distributions. After assuming that \( F_1 \) and \( F_2 \) are Gaussian with respective estimated mean vectors $\mu_1, \mu_2$ and covariance matrices $\Sigma_1,\Sigma_2$, this distance becomes
\begin{align}
\begin{split}
\label{eq:frechet}
d_F(F_1, F_2) &= \left(||\mu_1 - \mu_2||_2^2\right. \\
&\left. + \mathrm{tr}\left[\Sigma_1 + \Sigma_2 -2(\Sigma_1\Sigma_2)^{\frac{1}{2}} \right] \right)^{\frac{1}{2}}.
\end{split}
\end{align}

The popular \textbf{\Frechet{} Inception Distance (FID)} metric \citep{fid} computes this distance utilizing an ImageNet-pretrained \citep{imagenet} Inception v3 network \citep{szegedy2016rethinking} as the encoder. Other metrics include \textbf{KID (Kernel Inception Distance)} \citep{binkowski2018demystifying}, which uses Maximum Mean Discrepancy (MMD) and is suited for smaller datasets, and \textbf{CMMD (CLIP-MMD)} \citep{jayasumana2023rethinking}, which employs CLIP-extracted \citep{clip} image features with MMD as an alternative to FID. However, note that all of these distances utilize features learned from natural images.

\subsection{Radiology FID (RadFID)}
Recent studies suggest that standard perceptual metrics like FID, which are pretrained on natural images, may be unsuitable for medical images \citep{segdiff,konz2024rethinking,contourdiff,wu2025pragmatic}. A straightforward solution to this is given by the usage of features from a model trained on a large ``universal'' medical image dataset, such as Radiology ImageNet (RadImageNet) \citep{radimagenet}; such an approach, termed RadFID, was introduced in \citet{medigan} and then further tested for unconditional generative model evaluation in \citet{woodland2024_fid_med}. However, it has not seen widespread adoption, and this work is the first to explore its use for OOD detection as well as image translation.

\subsection{\Frechet{} Radiomic Distance (FRD): Version 0}
We introduced a preliminary version of the \Frechet{} Radiomic Distance in \citep{frd} in the context of evaluating multi-condition latent diffusion models for breast MRI generation, which we label here as ``FRD$_\text{\textbf{v0}}$''. FRD$_\text{\textbf{v0}}$ is computed by extracting 94 different radiomic features $v_{ji}$ for each image $x_i$ in the given dataset $D$ ($D_1$ or $D_2$), and min-max normalizing each type of feature (to $[0,1]$) given its distribution in the dataset, as
\begin{equation}
    \label{eq:minmaxnorm}
    v_{ji} \leftarrow \frac{v_{ji} - \mathrm{min}_iv_{ji}}{\mathrm{max}_iv_{ji} - \mathrm{min}_iv_{ji}}.
\end{equation}
Next, to make FRD$_\text{\textbf{v0}}$ values comparable to FID, these are re-scaled to the common range of ImageNet pre-trained InceptionV3 features used to compute FID, $[0, 7.456]$. Finally, FRD$_\text{\textbf{v0}}$ is computed as the \Frechet{} distance between these min-max normalized distributions of radiomic features for $D_1$ and $D_2$.

FRD makes several improvements to FRD$_\text{\textbf{v0}}$ (Sec. \ref{sec:radiomicdist}), including (1) the addition of many radiomic features that better capture the variation and nuances of different signals (\eg, in frequency space) that are useful to compare imaging distributions; (2) using $z$-score normalization instead of min-max for better robustness to outliers; and (3) normalizing a given feature type for both $D_1$ and $D_2$ with respect to the same reference distribution. Our experiments in using these metrics for a wide range of applications (Sec. \ref{sec:experiments}) show that these changes result in universal improvements.

\subsection{Evaluating Deep Generative Medical Image Models} We will study two types of generative models in medical imaging: image-to-image translation models and unconditional generative models.
Image-to-image translation models, primarily used in radiology to mitigate domain shift between datasets \citep{armanious2020medgan, mcnaughton2023machine}, have applications such as inter-scanner translation (e.g., across different manufacturers) \citep{cao2023deep, beizaee2023harmonizing}, inter-sequence translation (e.g., T1 to T2 MRI) \citep{durrer2024diffusion, li2023zero}, and inter-modality translation (e.g., MRI to CT) \citep{yang2019unsupervised,maskgan,zhang2018translating,wang2024mutual}. Our study includes datasets covering each of these scenarios.

Unconditional generative models, which learn to generate synthetic images given unlabeled real images for training, are commonly used to supplement medical image datasets, \eg, for the training of diagnostic models \citep{kazeminia2020gans, yi2019generative,chen2022generative}, including the generation of rare cases \citep{chen2021synthetic}.
To our knowledge, no previous work has developed task-independent metrics specifically for medical image generation or translation models. The vast majority of works utilize FID \citep{fid} (rather than \eg, RadFID), despite its aforementioned limitations.

\subsection{Radiomic Features for Medical Image Analysis} 
Radiomic features, which are typically hand-crafted, have long been used in diverse medical image diagnostic tasks \citep{pyradiomics,yip2016applications,gillies2016radiomics,lambin2012radiomics}, providing a meaningful, interpretable feature space for analyzing medical images. Applications include cancer screening \citep{jiang2021artificial}, outcome prediction \citep{aerts2014decoding,clark2008prognostic}, treatment response assessment \citep{drukker2018most,cha2017bladder,li2016mr}, and many others. A number of radiomic-based clinical tests have even received FDA clearance \citep{huang2023criteria}.
While previous works mainly use learned network features over pre-defined radiomic features for diagnostics \citep{wagner2021radiomics}, few studies have applied radiomics for out-of-domain detection or the evaluation of image translation/generation models, which we show has strong potential.

\section{Methods}
\label{sec:methods}

\subsection{Towards a Metric Designed for Medical Images: FRD}
\label{sec:radiomicdist}

\begin{figure*}[htbp!]
    \centering
    \includegraphics[width=0.99\linewidth]{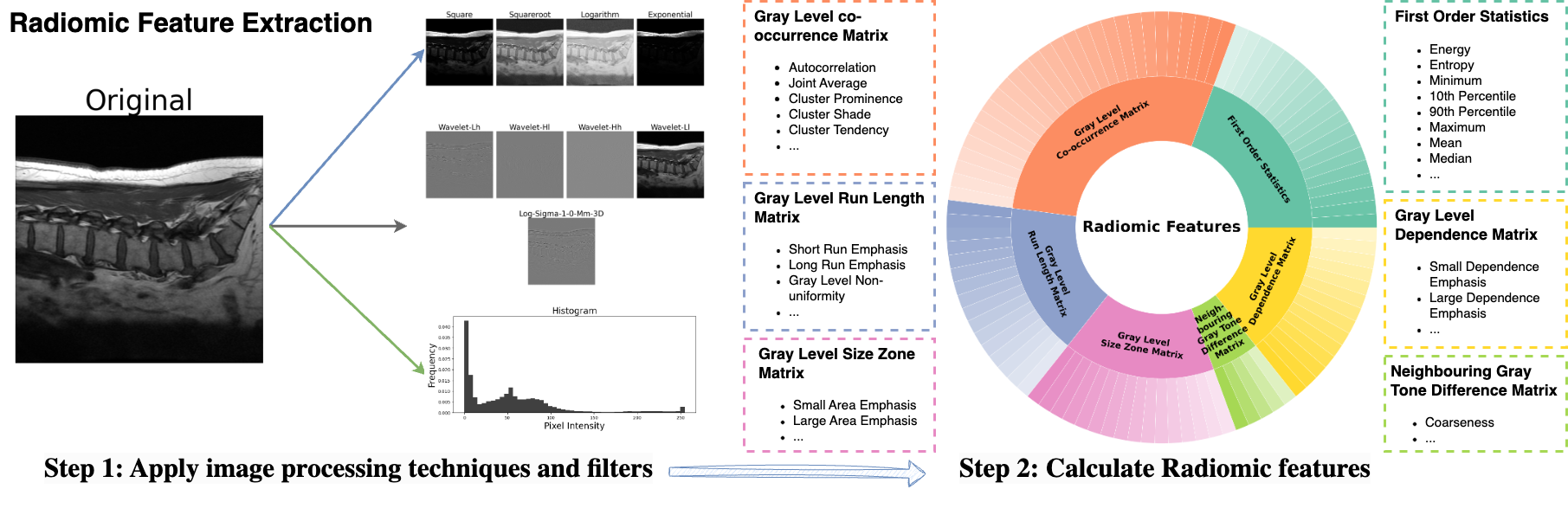}
    \caption{Our extraction process and taxonomy for radiomic image features. From left to right, the steps of feature extraction include \textbf{(1)} the passing of the input image through possible filters, \eg, wavelet/frequency-space conversion, and \textbf{(2)} the measurement of various first-order or higher-order features, \eg, entropy and gray-level matrix features, respectively.}
    \label{fig:radiomics_taxonomy}
\end{figure*}

RadFID is a seemingly suitable alternative to FID, which we will show improves on typically-used perceptual metrics for medical images---such as FID---in various aspects. However, it still lacks clear interpretability of the features being used to compare images (as well as various other limitations which we will demonstrate with upcoming experiments). For medical imaging, especially in image-to-image translation tasks, it is often critical to answer specific questions about how an image's features change or differ from some reference/original image---a need less relevant in natural image translation tasks like style transfer. However, the learned features used in RadFID/FID are difficult to interpret reliably (Sec. \ref{sec:interp}).

As a more interpretable alternative, we propose FRD, which utilizes a space of real-valued \textit{radiomic features} of images. The taxonomy and extraction process of these features are illustrated in Fig. \ref{fig:radiomics_taxonomy}. They include image-level features such as basic first-order statistics, and textural statistics such as the gray level co-occurrence matrix \citep{haralick1973textural_glcm}, gray level run length matrix \citep{galloway1975texture}, and gray level size zone matrix \citep{Thibault2009TextureIA}. Crucially, we improve on FRD$_\text{\textbf{v0}}$ \citep{frd} by passing input images through various optional wavelet filters prior to radiomic computation: these filters first apply a spatial Fourier transform to an image, and then apply one of four different possible choices of low- or high-pass filter combinations along the two spatial directions (\eg, low-low, low-high, high-low, and high-high).

This step results in a much more comprehensive space of features with which to compare distributions of images---in particular, the frequency-based features which can capture crucial subtleties in images---which we will show results in improved performance for a wide range of applications (Sec. \ref{sec:experiments}, \ref{sec:app:featureimportance}, etc.). In total, this results in $m = 464$ features, for each combination of filter and radiomic feature. All computations are completed via the PyRadiomics library \citep{pyradiomics}, and radiomics are extracted from the entire, unmasked image. Note that we evaluate the importance of different types of features for FRD in \ref{sec:app:featureimportance}. Importantly, these features are in compliance with definitions provided by the Imaging Biomarker Standardization Initiative \citep{zwanenburg2016image,zwanenburg2020image}, mitigating past issues of poorly-standardized radiomics in the field \citep{yip2016applications,METRICs}.

Each image \( x \in \mathbb{R}^n \) is mapped to its radiomic feature representation \( f_{\mathrm{radio}}(x) \in \mathbb{R}^m \), and we compute FRD as the \Frechet{} distance between radiomic feature distributions \( D_1 \) and \( D_2 \), applying a logarithmic transformation for stability:
\begin{equation}
    \label{eq:radiomicdist}
    \text{FRD}(D_1, D_2) := \log d_F(f_{\mathrm{radio}}(D_1), f_{\mathrm{radio}}(D_2)).
\end{equation}
We also $z$-score normalize each feature with respect to its distribution in \( D_1 \). Note that we tested MMD distance as an alternative to \Frechet{}, but found it less effective (\ref{sec:app:mmd}).

\subsection{Downstream Task-Based Image Metrics}
\label{sec:downstreamtasks}
FRD and the other perceptual metrics discussed so far compare image distributions in a downstream task-independent manner. However, in medical image analysis, task-dependent metrics are often more common, such as assessing how closely translated/generated images resemble real target images by evaluating them with downstream tasks such as semantic segmentation. For example, in image translation, this often involves training a model on real target domain images and testing it on translated images \citep{vorontsov2022towards,kang2023structure}, or vice versa \citep{yang2019unsupervised,contourdiff}. If the task is segmentation, this approach also measures \textit{anatomical consistency} between source and translated images. Such metrics will therefore serve as important targets for the task-independent metrics which we will evaluate.

While segmentation is the primary task of interest, we also assess other downstream tasks including object detection and classification. We denote downstream performance on a dataset \( D \) (\eg, a translated image test set \( \Dst^\mathrm{test} \)) as \( \mathrm{Perf}(D) \in \mathbb{R} \), with higher values indicating better performance. Specifically, we use the Dice coefficient for segmentation, mIoU (mean Intersection-over-Union) and mAP@[$0.5, 0.95$] (mean Average Precision) for object detection, and AUC (area under the receiver operating characteristic curve) for classification---the latter computed on predicted logits to account for test set class imbalance \citep{mandrekar2010receiver,maier2024metrics}.

\subsection{Datasets and Downstream Tasks}
\label{sec:datasets}

\begin{figure}[htbp!]
    \centering
    \includegraphics[width=0.95\linewidth]{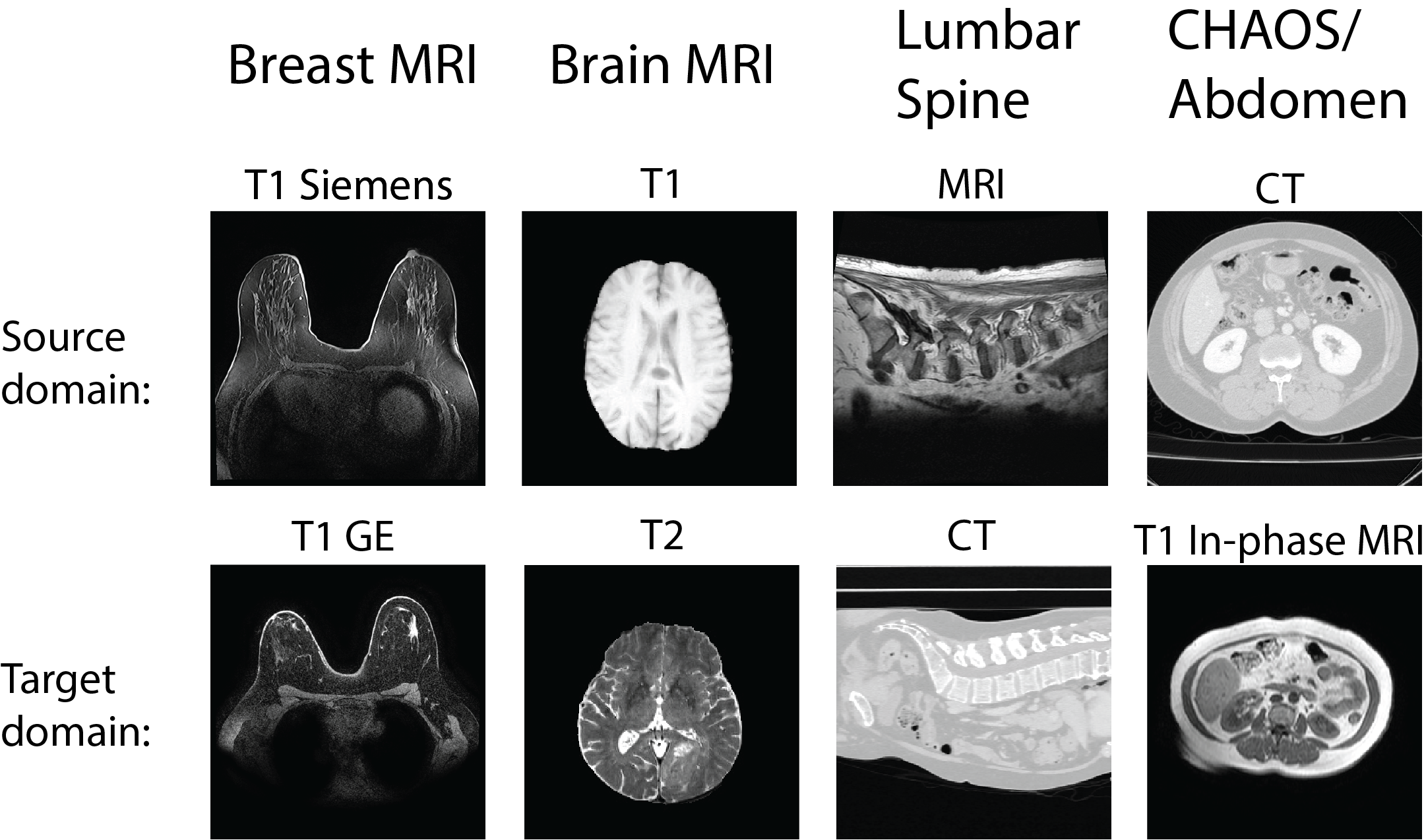}
    \caption{Example images from each dataset, ordered left-to-right with respect to Table \ref{tab:datasets}.}
    \label{fig:datasets}
\end{figure}

We evaluate a range of multi-domain medical (radiology) datasets of 2D image slices extracted from 3D volumes for out-of-domain detection, translation, and generation, covering inter-scanner, inter-sequence, and inter-modality cases (ordered from least to most severe visual differences between domains). The datasets include: \textbf{(1) breast MRI} (T1-weighted) from Siemens and GE scanners (DBC \citep{saha2018machine,lew2024publicly}); \textbf{(2) brain MRI} (T1-weighted and T2-weighted sequences) from BraTS \citep{brats}; \textbf{(3) lumbar spine} MRIs and CTs (from TotalSegmentator \citep{Wasserthal_2023} and in-house MRIs); and \textbf{(4) abdominal CT and in-phase T1 MRI} from CHAOS \citep{kavur2021chaos}. Each dataset is split by patient into training, validation, and test sets, which are each sub-split into domains (details in Table \ref{tab:datasets}), resized to \(256 \times 256\) and normalized to \([0,1]\). Example images are in Fig. \ref{fig:datasets}.

\begin{table*}[htbp!]
  \centering
  \fontsize{8pt}{8pt}\selectfont
  \begin{tabular}{lllllcl}
    \toprule
    \bf Abbrev. name & \bf Full name/citation & \bf Domains & \bf Intra- & \bf Inter- & \bf Train/val/test sizes & \bf Downstream tasks \\\midrule
    Breast MRI & \tworow{Duke Breast Cancer}{\tworow{MRI (DBC)}{\tworow{\citep{saha2018machine},}{\citep{lew2024publicly}}}} & \tworow{Siemens$\rightarrow$GE}{(T1 MRI)} & Sequence & \tworow{Scanner}{Manuf.} & 12K/2.4K/2.6K & \tworow{\tworow{FGT segmentation,}{breast segmentation,}}{cancer classification} \\\midrule
    Brain MRI & \tworow{BraTS}{\citep{brats}} & T1$\rightarrow$T2 & Modality & Sequence & 28K/6K/6K & \tworow{\tworow{Tumor segmentation,}{tumor detection,}}{cancer classification} \\\midrule
    Lumbar spine & \tworow{\tworow{TotalSegmentator}{\citep{Wasserthal_2023}}}{and in-house MRIs} & T1 MRI$\rightarrow$CT & Body region & Modality & 2K/0.6K/0.6K & Bone segmentation \\\midrule
    CHAOS & \tworow{\tworow{CHAOS}{(Abdom. MRI \& CT)}}{\citep{kavur2021chaos}} & \tworow{CT$\rightarrow$T1 MRI}{(in-phase)} & Body region & Modality & 1.8K/1.1K/0.6K & \tworow{Liver segmentation,}{liver classification} \\
    \bottomrule
\end{tabular}
\caption{\textbf{Main datasets evaluated in this paper.} ``Domains'' are the source$\rightarrow$target domain pairs used, \eg, for image translation. Each dataset is labeled for how its domains are similar (``Intra-'') and how they differ (``Inter-''), \eg, the BraTS dataset has intra-modality, inter-sequence domains. ``FGT'' is fibroglandular tissue.}
\label{tab:datasets}
\end{table*}

The lumbar spine and CHAOS datasets pose especially challenging scenarios due to their relatively small size and significant differences in visible features and anatomical structures between their respective domains.

\paragraph{Downstream Task Evaluation}
In addition to prior task-independent/perceptual metrics (Sec. \ref{sec:relwork}) and FRD (Sec. \ref{sec:radiomicdist}), we assess images using auxiliary models trained on downstream tasks (Table \ref{tab:datasets}), as described in Sec. \ref{sec:downstreamtasks}. These models, trained on target domain data, are tested on various domains, such as target (\(\Dt^\mathrm{test}\)), source (\(\Ds^\mathrm{test}\)), source-to-target translations (\(\Dst^\mathrm{test}\)), or others, depending on the experiment. Full details on model training, architecture, dataset creation, and task labels are provided in \ref{app:training:downstream} and \ref{app:datasets}.

\section{Evaluation and Results}
\label{sec:experiments}

We will now demonstrate various applications of FRD, including out-of-domain detection (Sec. \ref{sec:ood}), image-to-image translation evaluation (Sec. \ref{sec:mainexp}), unconditional image generation evaluation (Sec. \ref{sec:uncondmodels}), and abnormality detection (Sec. \ref{sec:FRDpredictor}).

\subsection{FRD for Out-of-Domain Detection}
\label{sec:ood}

As discussed in the introduction, a common problem in deep learning for medical image analysis is \textit{domain shift}: where when some diagnostic downstream task model is presented with images that were acquired from a site, sequence, or modality different from the one where its training data originated, there may be a performance drop due to the data being OOD from the training data \citep{albadawy2018deep,maartensson2020reliability,guo2024impact}. In this section, we will show how FRD is overall superior to prior perceptual metrics for detecting when medical images are OOD.

Perceptual metrics like FRD and FID can help detect whether a new image \( x_{\text{test}} \) is in-distribution (ID) or out-of-distribution (OOD) relative to a reference ID dataset \( D_\mathrm{ID} \) (\eg, some model's training set) without labels. The OOD score \( s(x_{\text{test}};D_\mathrm{ID}) \) can be defined as the distance (here, we use $L_2$) of \( x_{\text{test}} \)'s features from the mean features of \( D_\mathrm{ID} \):
\begin{equation}
    \label{eq:ood}
    s(x_{\text{test}};D_\mathrm{ID}) = ||f(x_{\text{test}}) - \mathbb{E}_{x_\mathrm{ID} \sim D_\mathrm{ID}} f(x_\mathrm{ID})||_2,
\end{equation}
where \( f \) is an image feature encoder \citep{reiss2023mean, schlegl2019f}. OOD performance thus depends on the choice of feature space, so we compare radiomic (\ie, FRD or FRD$_\text{\textbf{v0}}$), RadImageNet (\ie, RadFID) and ImageNet (\ie, FID) features for OOD detection.

\begin{figure}[htbp!]
    \centering
    \includegraphics[width=0.99\linewidth]{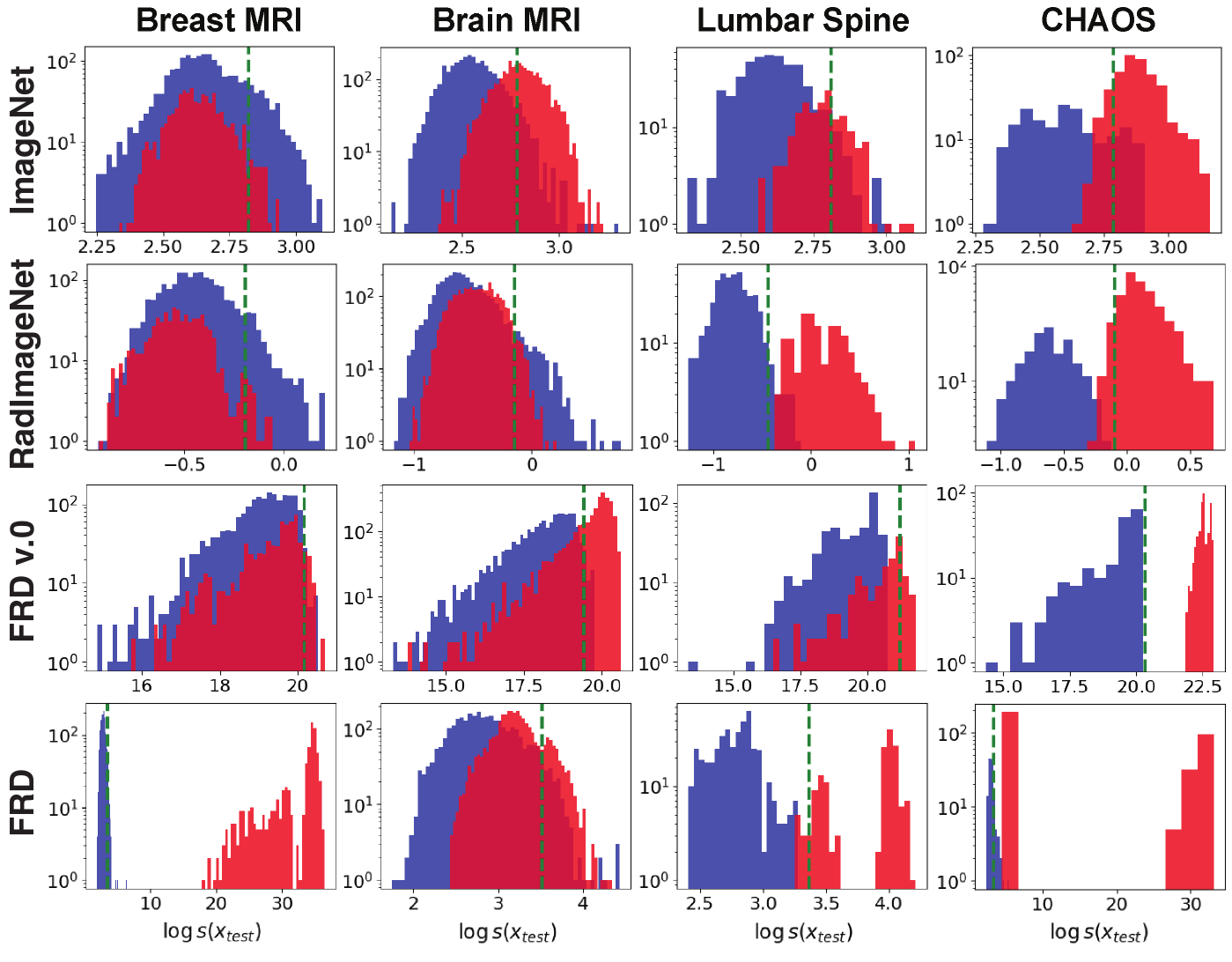}
    \caption{OOD detection score distributions for in-domain (\textcolor{blue}{\textbf{blue}}) and OOD (\textcolor{red}{\textbf{red}}) test images for each dataset (columns), using different feature spaces (rows). Computed detection thresholds (Eq. \ref{eq:ood_thresh}) are shown as vertical dashed \textcolor{OliveGreen}{\textbf{green}} lines.}
    \label{fig:ood_dists}
\end{figure}

For each dataset, we use the target domain training set (Table \ref{tab:datasets}) as \( D_\mathrm{ID} \) and compute the OOD score on the ID and OOD images of the test set\footnote{Note that for datasets which have images from multiple domains of the same patient, \eg, BraTS, we use random sampling to ensure that the domain subsets for a given train/val/test set do not overlap by patient.}, aggregating scores via AUC \citep{fawcett2006introduction}. OOD detection AUC results (top block of Table \ref{tab:ood}) and ID vs. OOD detection score distributions (Fig. \ref{fig:ood_dists}) show that FRD radiomic features outperform learned feature spaces (FID, RadFID) and FRD$_\text{\textbf{v0}}$ radiomic features on average, more clearly separating ID and OOD distributions. This is particularly true for the challenging case of breast MRI, where the domain shift is visually subtle (as shown in Fig. \ref{fig:datasets}), likely due to the use of frequency-space features in FRD, which can capture subtle visual details.

\begin{table}[htbp]
\setlength{\tabcolsep}{2pt}
\fontsize{8pt}{8pt}\selectfont
\centering
\begin{tabular}{l|l|cccc||c}
\textbf{\tworow{}{Metric}} & \textbf{\tworow{Feature Space/}{Distance Metric}} & \textbf{\tworow{Breast}{MRI}} & \textbf{\tworow{Brain}{ MRI}} & \textbf{\tworow{}{Lumbar}} & \textbf{\tworow{}{CHAOS}} & \textbf{\tworow{}{Avg.}} \\
\midrule
\multirow{4}{*}{\textbf{AUC}} & ImageNet    & 0.43       & \textbf{0.91}      & 0.89   & 0.94  & 0.79 \\
 & RadImageNet & 0.35      & 0.64      & \underline{0.99}   & \underline{0.99}  & 0.74 \\
 & FRD$_\text{\textbf{v0}}$ & \underline{0.60} & \underline{0.90} & 0.78 & \textbf{1.00} & \underline{0.82} \\
 & FRD   & \textbf{1.00}       & 0.76      & \textbf{1.00}   & \textbf{1.00}  & \textbf{0.94} \\\midrule
\multirow{4}{*}{\textbf{Accuracy}} & ImageNet    & 0.65       & \underline{0.73}     & 0.81   & 0.84  & 0.76 \\
 & RadImageNet & 0.68       & 0.48      & \textbf{0.98}   & 0.92  & 0.77 \\
 & FRD$_\text{\textbf{v0}}$ & \underline{0.74} & \textbf{0.84} & 0.79 & \textbf{1.0} & \underline{0.84} \\
 & FRD   & \textbf{0.96}      & 0.57      & \underline{0.92}   & \underline{0.95}  & \textbf{0.85} \\\midrule
\multirow{4}{*}{\textbf{Sensitivity}} & ImageNet    & 0.03       & 0.51      & 0.37   & 0.83  & 0.44 \\
 & RadImageNet & 0.02       & 0.03      & \textbf{1.00}   & \underline{0.88}  & 0.48 \\
 & FRD$_\text{\textbf{v0}}$ & \underline{0.07} & \underline{0.71} & 0.19 & \textbf{1.00} & \underline{0.49} \\
 & FRD   & \textbf{1.00}       & \textbf{0.95}      & \underline{0.71}   & \textbf{1.00}  & \textbf{0.92} \\\midrule
\multirow{4}{*}{\textbf{Specificity}} & ImageNet    & 0.88       & \underline{0.95}      & 0.96   & \underline{0.87}  & 0.92 \\
 & RadImageNet & 0.92   & 0.94      & \underline{0.97}   & \textbf{1.00}  & \underline{0.96} \\
 & FRD$_\text{\textbf{v0}}$ & \textbf{0.99} & \textbf{0.97} & \textbf{1.00} & \textbf{1.00} & \textbf{0.99} \\
 & FRD   & \underline{0.95}     & 0.92      & \textbf{1.00}   & 0.85  & 0.93 \\
\bottomrule
\end{tabular}
\caption{Using different feature spaces/distance metrics for OOD detection. Best result and runner-up for a given detection metric and dataset are shown in bold and underlined, respectively. 95th percentile thresholding (Eq. \ref{eq:ood_thresh}) was used to obtain accuracy, sensitivity and specificity results.}
\label{tab:ood}
\end{table}

However, for true practical use, a score threshold \( \hat{s} \) would need to be set to binarily classify ID vs. OOD images \textit{without} a validation set of OOD examples, as AUC simply integrates over all possible thresholds. Given the lack of OOD examples, this is doable if we heuristically
set \( \hat{s} \) as the 95th percentile of the scores for known ID points:
\begin{equation}
    \label{eq:ood_thresh}
    \hat{s} = \mathrm{Percentile}_{95}(S_{\mathrm{ID}}),
\end{equation}
where $S_{\mathrm{ID}}$ is the reference distribution of ID scores defined by $S_\mathrm{ID} := \{s(x;D_\mathrm{ID} \setminus x) : x \in D_\mathrm{ID}\}$.
We illustrate these computed thresholds for each dataset using different feature spaces in Fig. \ref{fig:ood_dists}, and show quantitative detection results using them in Table \ref{tab:ood}. We see that using FRD or FRD$_\text{\textbf{v0}}$ features over ImageNet or RadImageNet results in noticeably improved average accuracy and sensitivity, and on-par specificity, especially for the challenging subtle domain shift case of breast MRI. FRD improves on FRD$_\text{\textbf{v0}}$ noticeably in AUC and sensitivity, and is roughly on-par for accuracy and specificity.

\subsubsection{OOD Performance Drop Prediction}
Another closely related question is \textit{``does FRD detect when performance will drop on new data?''} for some downstream task model. We evaluated this for each of the downstream tasks of Table \ref{tab:datasets}, and we see that in almost all cases, there is a drop in average performance on test data that was detected as OOD using the binary threshold approach of Eq. \ref{eq:ood_thresh}, compared to ID performance (full table in \ref{app:results_tasksOODdrop}). Additionally, in \ref{app:ood_drop_full} we show that FRD outperforms other metrics in ranking which of different OOD datasets will result in worse downstream task performance.

\subsubsection{Towards Practical Dataset-Level OOD Detection}
We also propose a FRD-based metric for \textit{dataset-level} OOD detection, $\text{nFRD}_\mathrm{group}$, which is formulated to estimate the probability that some new test set $D_\mathrm{test}$ is OOD as a whole, relative to $D_\mathrm{ID}$.
This is particularly designed for the realistic scenario of receiving a new dataset from some outside hospital/site, and wanting an interpretable indication of if the dataset is suitable for some in-domain trained model. While preliminary, we found that this metric scores OOD datasets more consistently than other prior metrics, providing an estimated OOD probability of $\text{nFRD}_\mathrm{group}\simeq 1$ for 3 out of 4 datasets which span a variety of modalities and body regions; see \ref{app:ood_datasetlevel} for the full details.

\subsubsection{Summary: Practical Medical Image OOD Detection using FRD}
Finally, in the interest of practical usage, we provide a step-by-step guide for OOD detection of medical images using FRD in Algorithm \ref{alg:ood_detect}. 

\begin{algorithm}[htbp!]
    \caption{Medical Image OOD Detection using FRD.}
    \label{alg:ood_detect}
    \small
\begin{algorithmic}[1] 
    \REQUIRE{Test image set $D_\mathrm{test}$, reference ID image set $D_\mathrm{ID}$, radiomic feature encoder $f:=f_\mathrm{radio}$.}
    \STATE $S_\mathrm{ID} := \{s(x_\mathrm{ID};D_\mathrm{ID} \setminus x_\mathrm{ID}) : x_\mathrm{ID} \in D_\mathrm{ID}\}$
    \STATE $\hat{s} = \mathrm{Percentile}_{95}(S_{\mathrm{ID}})$
    \STATE $\ell_\mathrm{test} := \{\mathbf{1}[s(x_\mathrm{test};D_\mathrm{ID}) \geq \hat{s}]  : x_\mathrm{test} \in D_\mathrm{test}\}$
    \RETURN Binary OOD labels $\ell_\mathrm{test}$
    \RETURN (Optional) dataset-level OOD score, $\text{nFRD}_{\mathrm{group}}(D_\mathrm{test};D_\mathrm{ID})$ (\ref{app:ood_datasetlevel})
\end{algorithmic}
\end{algorithm}

\subsection{FRD for Evaluating Image-to-Image Translation}
\label{sec:mainexp}


\subsubsection{Image-to-Image Translation Models}
\label{sec:models}

Unpaired image-to-image translation for medical images, lacking paired data, is challenging and typically relies on adversarial learning. We evaluate a variety of state-of-the-art unpaired models: CycleGAN \citep{cyclegan}, MUNIT \citep{munit}, CUT \citep{cut}, GcGAN \citep{gcgan}, MaskGAN \citep{maskgan}, and UNSB \citep{unsb}, each representing diverse techniques such as contrastive learning and style/content disentanglement. All models are trained on source and target domain images from each dataset, with detailed training specifics in \ref{app:training:trans}.

\subsubsection{Evaluation with Perceptual Metrics}
\label{sec:exp:perceptual}

\begin{figure}[thpb]
    \centering
    \includegraphics[width=0.99\linewidth]{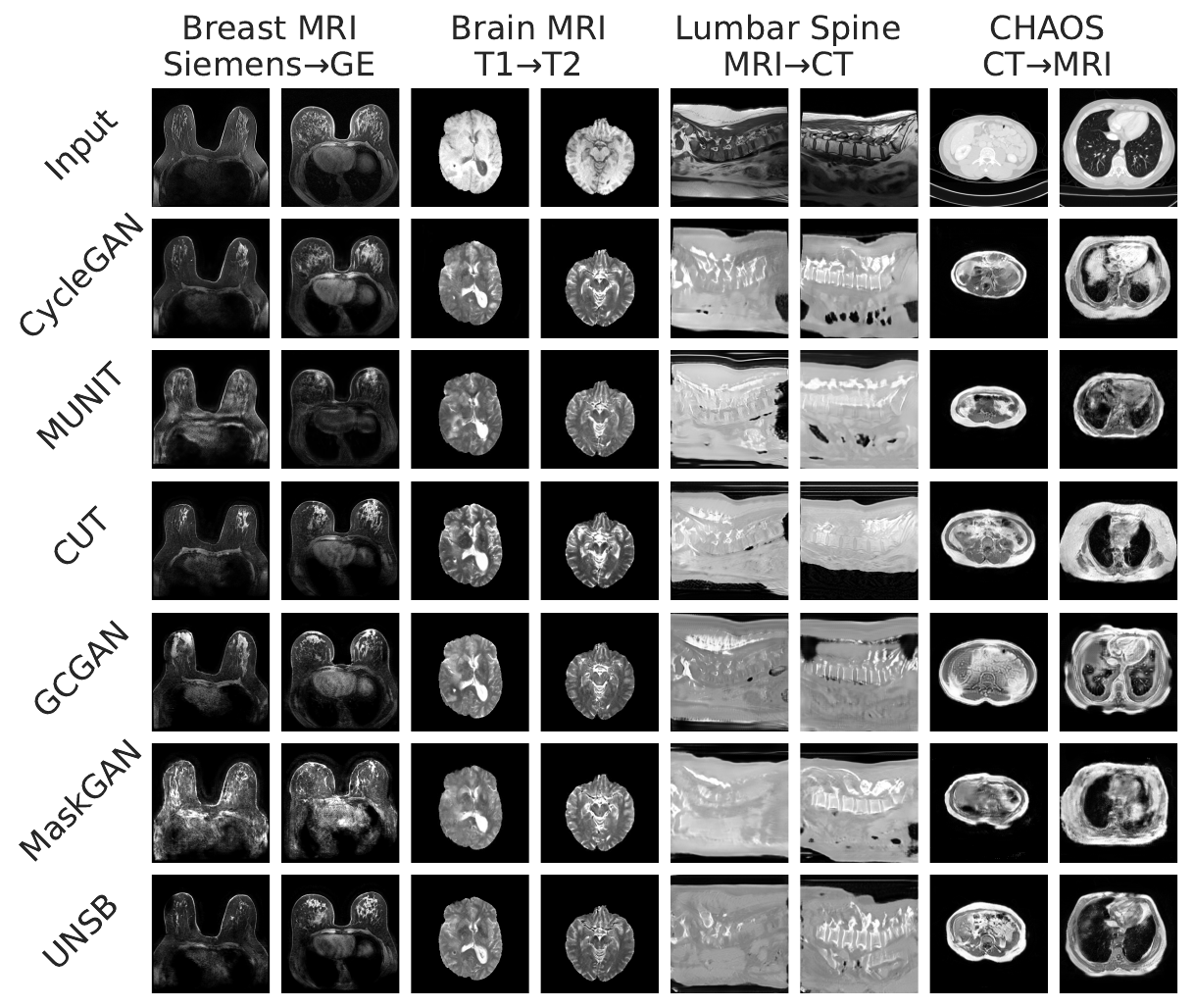}
    \caption{Translations $x_{s\rightarrow t}^\mathrm{test}$ from each translation model (non-top rows) given example inputs $x_s^\mathrm{test}$ (top row).}
    \label{fig:translated_examples}
\end{figure}

We first evaluate each translation model using perceptual metrics to measure the distance between translated test set source domain images and real test set target domain images. We compare FRD to FRD$_\text{\textbf{v0}}$, RadFID, FID, KID, and CMMD for this task, with results in Table \ref{tab:results_perceptual}. 
We first qualitatively observe that FID often fails to capture visual quality and anatomical consistency, particularly when there is a high semantic shift between source and target domains, as shown in Fig. \ref{fig:translated_examples}. For example, FID, KID and CMMD rate MUNIT as best for lumbar spine despite a clear loss of bone structure---shown by MUNIT being the \textit{worst} by segmentation performance in Table \ref{tab:results_tasks}, which FRD and RadFID capture successfully. 
This highlights certain limitations of using prior perceptual metrics for medical images.

\begin{table*}[htbp]
    \setlength{\tabcolsep}{2pt}
    \centering
    \fontsize{8pt}{8pt}\selectfont

    \begin{tabular}{l|cccccc||cccccc}
        \multicolumn{1}{c}{} & \multicolumn{6}{c||}{\textbf{Breast MRI}} & \multicolumn{6}{c}{\textbf{Brain MRI}} \\
        \toprule
        \textbf{Method} & FRD & FRD$_\text{\textbf{v0}}$ & RadFID & FID & KID  & CMMD  & FRD & FRD$_\text{\textbf{v0}}$ & RadFID & FID  & KID & CMMD  \\
        \midrule
        CycleGAN & 38.1 & 706  & 0.26 & 107 & 0.049 & \textbf{0.308} & 33.4 & 554  & 0.06 & 21.7 & \underline{0.004} & 0.378 \\
        MUNIT & 43.7 & \textbf{626}  & 0.29 & 144 & 0.089 & 1.480 & \underline{25.7} & 540  & \underline{0.05} & \underline{21.6} & 0.006 & 0.388 \\
        CUT & \underline{24.8} & \underline{632}  & \textbf{0.17} & 106 & 0.053 & 0.362 & 33.8 & 552  & 0.13 & 29.4 & 0.012 & \underline{0.259} \\
        GcGAN & \textbf{24.6} & 647  & \textbf{0.17} & \underline{104} & \underline{0.040} & \underline{0.322} & \textbf{12.1} & \textbf{523}  & \textbf{0.04} & \textbf{19.0} & \textbf{0.003} & \textbf{0.239} \\
        MaskGAN & 48.7 & 1042  & 0.35 & 118 & 0.089 & 0.642 & 27.8 & 555  & 0.06 & 23.5 & 0.008 & 0.392 \\
        UNSB & \textbf{24.6} & 645  & \underline{0.19} & \textbf{91} & \textbf{0.033} & 0.388 & \textbf{12.1} & \underline{525}  & 0.08 & 26.0 & 0.010 & 0.563 \\
        \bottomrule
    \end{tabular}
    
    \vspace{1em} 
    
    \begin{tabular}{l|cccccc||cccccc}
        \multicolumn{1}{c}{} & \multicolumn{6}{c||}{\textbf{Lumbar}} & \multicolumn{6}{c}{\textbf{CHAOS}} \\
        \toprule
        \textbf{Method} & FRD & FRD$_\text{\textbf{v0}}$ & RadFID & FID & KID  & CMMD  & FRD & FRD$_\text{\textbf{v0}}$ & RadFID & FID  & KID & CMMD  \\
        \midrule
        CycleGAN & 6.71 & \underline{350}  & 0.25 & 210 & \underline{0.161} & 2.950 & 42.8 & 470  & \underline{0.11} & \textbf{122} & \textbf{0.051} & \underline{0.379} \\
        MUNIT & 9.31 & 367  & 0.30 & \textbf{197} & \textbf{0.151} & \textbf{2.317} & \textbf{5.41} & \textbf{276}  & \textbf{0.10} & 136 & 0.073 & 0.904 \\
        CUT & \textbf{6.48} & 417  & \textbf{0.21} & 245 & 0.206 & 3.373 & 6.84 & 514  & \textbf{0.10} & 145 & 0.083 & 0.444 \\
        GcGAN & \underline{6.52} & \textbf{313}  & 0.25 & 226 & \underline{0.161} & 3.300 & \underline{6.38} & \underline{434}  & 0.12 & 141 & \underline{0.064} & 0.507 \\
        MaskGAN & 6.64 & 421  & 0.27 & 248 & 0.217 & 3.237 & 58.8 & 437  & 0.22 & 212 & 0.130 & 2.120 \\
        UNSB & 6.59 & 375  & \underline{0.23} & \underline{208} & 0.172 & \underline{2.579} & 51.8 & 542  & \underline{0.11} & \underline{135} & 0.078 & \textbf{0.356} \\
        \bottomrule
    \end{tabular}
    
    \caption{
    \textbf{Perceptual/task-independent metrics $d(\Dst^\mathrm{test}, \Dt^\mathrm{test})$ for image translation models.} Best and runner-up models according to each metric in bold and underlined, respectively.}
    \label{tab:results_perceptual}
\end{table*}

\subsubsection{Correlation with Downstream Task Performance and Anatomical Consistency}
\label{sec:exp:downstreamcorr}

Since a key goal of medical image translation is maintaining downstream task performance (\eg, segmentation) and mitigating domain shift, we will now examine whether perceptual metrics can serve as proxies for task performance by correlating perceptual distances with downstream task metrics. We calculate the Pearson correlation \( r \) between each perceptual metric and downstream performance across all translation models (Tables \ref{tab:results_perceptual} and \ref{tab:results_tasks}).

\begin{table*}[htbp]
    \setlength{\tabcolsep}{2pt}
    \centering
    \fontsize{8pt}{8pt}\selectfont
    \begin{tabular}{l|cc|c||c|cc|c||c||cc}
    \multicolumn{1}{c}{} & \multicolumn{3}{c||}{\textbf{Breast MRI}} & \multicolumn{4}{c||}{\textbf{Brain MRI}} &  \multicolumn{1}{c||}{\textbf{Lumbar}} & \multicolumn{2}{c}{\textbf{CHAOS}}  \\
    \toprule
    \multicolumn{1}{c|}{} & \multicolumn{2}{c|}{Dice} &  AUC& \multicolumn{1}{c|}{Dice} & mIoU& mAP & AUC& \multicolumn{1}{c||}{Dice} & \multicolumn{1}{c}{Dice} & AUC\\
    \midrule 
    \textbf{Method} & Breast & FGT & \multicolumn{1}{c|}{Cancer} & Tumor & \multicolumn{2}{c|}{Tumor}  & Cancer & Bone & Liver & Liver  \\
    \midrule
    CycleGAN  & 0.871 & \textbf{0.494} & 0.530 & 0.348 & 0.164 & 0.126 & 0.805  & \underline{0.232} & 0.284 & 0.591 \\
    MUNIT  & 0.832 & 0.201  & 0.511 & 0.337 & 0.168 & 0.125 & \underline{0.844}  & 0.101 & 0.182 & 0.323 \\
    CUT  & 0.843 & 0.373  & \underline{0.544} & 0.303 & 0.159 & 0.133 & \textbf{0.861}  & \textbf{0.277} & \textbf{0.444} & \textbf{0.744} \\
    GcGAN & \underline{0.876} & \underline{0.389} & 0.492 & \underline{0.360} & 0.165 & \textbf{0.137} & 0.835 & 0.126 & 0.167 & \underline{0.702} \\
    MaskGAN  & 0.809 & 0.164  & 0.441 & \textbf{0.375} & \textbf{0.170} & 0.126 & 0.842  & 0.167 & 0.317 & 0.375 \\
    UNSB  & \textbf{0.881} & 0.308  & \textbf{0.594} & 0.353 & \underline{0.169} & \underline{0.135} & 0.839 & 0.138 & \underline{0.381} & 0.405 \\
    \midrule
    \rowcolor{gray!30}\textit{In-domain} & 0.883 & 0.696  & 0.670 & 0.442 & 0.174 & 0.169 & 0.841  & 0.949 & 0.864 & 0.866 \\
    \rowcolor{gray!30}\textit{OOD} & 0.747 & 0.446 & 0.538 & 0.005 & 0.152 & 0.065 & 0.727  & 0.007 & 0.062 & 0.504 \\
    \bottomrule
    \end{tabular}
    
    \caption{\textbf{Downstream task performance metrics $\mathrm{Perf}(\Dst^\mathrm{test})$ for image translation models.} In-domain and out-of-domain performance shown at the bottom for reference for how susceptible each task is to domain shift, and as \textit{expected} upper and lower performance bounds.}
    \label{tab:results_tasks}
\end{table*}

\begin{figure}[thpb]
    \centering
    \includegraphics[width=0.99\linewidth]{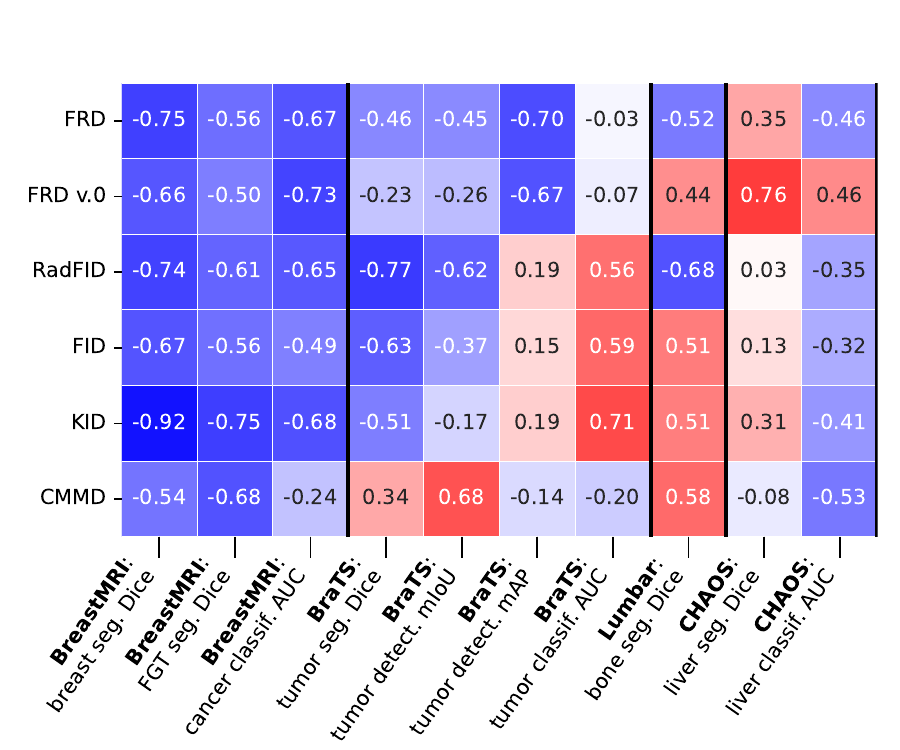}
    \caption{Pearson correlation of perceptual metrics (vertical axis) (Table \ref{tab:results_perceptual}) with downstream task-based metrics (horizontal axis) (Table \ref{tab:results_tasks}) for evaluating image translation, taken across all translation models \textbf{(lower $r$ (colder color) is better)}.}
    \label{fig:metric_corr}
\end{figure}

As shown in Fig. \ref{fig:metric_corr}, FRD has the strongest (most negative) average correlation with downstream task performance (\( r = -0.43 \)), followed by RadFID (\( r = -0.36 \)), while FRD$_\text{\textbf{v0}}$, FID, KID, and CMMD are less consistent ($r=-0.01$, $r=-0.17$, $r=-0.17$, $r=-0.08$, respectively), especially for datasets with larger domain shifts. This is particularly the case for segmentation tasks (which measure anatomical consistency), where---excluding CHAOS, which had low correlations likely due to it generally being a difficult dataset (see Sec. \ref{sec:datasets})---FRD and RadFID achieved mean correlations of $r=-0.58$ and $r=-0.70$, while FRD$_\text{\textbf{v0}}$, FID, KID and CMMD have $r=-0.04$, $r=-0.34$, $r=-0.42$, and $r=-0.08$, respectively.

These results align with recent work which showed that metrics like FID and KID, despite being popular in mainstream computer vision, do not reliably correlate with downstream task performance \citep{konz2024rethinking,wu2025pragmatic}. Our results also potentially indicate generally best translation models for medical images, which we will discuss in Sec. \ref{sec:discussion}.

\subsection{FRD for Evaluating Unconditional Image Generation}
\label{sec:uncondmodels}
We have so far focused our experimental effort on image-to-image translation over image generation due to the direct relationship of it with the key problem of domain shift in medical imaging.
We will now study FRD in evaluating unconditional generative medical image models, similar to FID's typical use. We trained StyleGAN2-ADA \citep{karras2020training} with default settings on four single-domain image generation tasks: \textbf{(1)} GE T1 breast MRI, \textbf{(2)} T1 brain MRI (BraTS), \textbf{(3)} lumbar spine CT, and \textbf{(4)} an abdominal CT dataset (CT-Organ \citep{rister2020ct})\footnote{CHAOS was not large enough to train on for high generation quality.}.

\begin{figure}[htbp!]
    \centering
    \setlength{\tabcolsep}{3pt}
    \fontsize{9pt}{9pt}\selectfont
    \begin{tabular}{l|cc|cc|cc|cc}
    \multicolumn{1}{c}{} & \multicolumn{2}{c|}{\textbf{\tworow{Breast}{MRI}}} & \multicolumn{2}{c|}{\textbf{\tworow{Brain}{MRI}}} &  \multicolumn{2}{c|}{\textbf{\tworow{Lumbar}{CT}}} & \multicolumn{2}{c}{\textbf{\tworow{Abdom.}{CT}}}  \\
    \toprule
    \textbf{Model:} & \textbf{A} & \textbf{B} & \textbf{A} & \textbf{B} & \textbf{A} & \textbf{B} & \textbf{A} & \textbf{B} \\
    \midrule
    FID & 67  & 54  & 157 & 26 & 210 & 103 & 232 & 118 \\
    KID & 0.02  & 0.01  & 0.14 & 0.01  & 0.18 & 0.04 & 0.23 & 0.08 \\
    CMMD & 0.30  & 0.33 & 0.82 & 0.28 & 3.57 & 0.87 & 4.72 & 0.48 \\
    RadFID & 0.16  & 0.06  & 0.16 & 0.04 & 0.20 & 0.11 & 0.40 & 0.16 \\
    FRD$_\text{\textbf{v0}}$ & 585 & 568 & 709 & 633 & 232 & 167 & 897 & 710 \\
    FRD & 47.4  & 43.4  & 58.5 & 56.3 & 8.3 & 6.3 & 71.3 & 68.6 \\
    \bottomrule
    \end{tabular}
    \vspace{0.4cm}
        
    \includegraphics[width=0.99\linewidth]{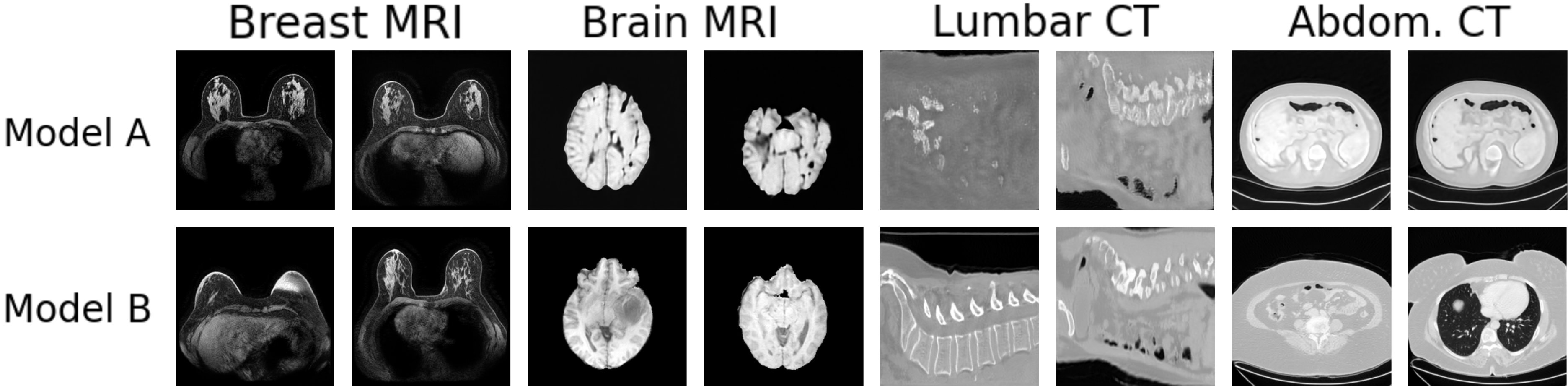}
    \caption{\textbf{Top: FRD and other perceptual metrics for evaluating unconditional generative models,} comparing a poor model (\textbf{A}) to a better model (\textbf{B}). \textbf{Bottom:} example generated images.}
    \label{fig:uncond_gen_combined}
\end{figure}

We evaluate each perceptual metric (FRD, FRD$_\text{\textbf{v0}}$, RadFID, FID, CMMD, KID) by ranking samples from an early model iteration (\textbf{Model A}, $2\times 10^5$ images seen in training) of visibly lower quality against a fully trained model (\textbf{Model B}, 10$\times$ more images seen in training)---similar to \citet{jayasumana2023rethinking}, with results and sample generated images shown in Fig. \ref{fig:uncond_gen_combined}. FRD successfully identifies the lower-quality model in all cases, aligning with prior metrics, except for CMMD, which fails for breast MRI.

\subsection{FRD as a Predictor of Abnormality}
\label{sec:FRDpredictor}
In this section, we will demonstrate a basic example of another use of FRD: as a predictor of abnormalities within medical images. We consider a dataset of axial breast DCE-MRI images cropped through the middle to only include one breast. We consider using FRD to predict whether a given test breast image is healthy or unhealthy in two scenarios: \textbf{(1)} where only a reference set of healthy breast images $D_\mathrm{healthy}$ is available (an anomaly detection/out-of-distribution detection scenario), and \textbf{(2)}, a somewhat easier case where a reference set of unhealthy breasts $D_\mathrm{cancer}$ can also be used. The test set and reference sets are sampled from the MAMA-MIA \citep{mamamia} test and train sets, respectively, with full dataset creation and pre-processing details provided in \ref{app:frdpredictordata}. 

In the first scenario, for each test image $x$, we measure the FRD of $x$ from the reference healthy set via $s(x; D_\mathrm{healthy})$ (Eq. \ref{eq:ood}), and use this as the unhealthy prediction score for $x$. We then aggregate all results via the AUC, computed with the $s(x; D_\mathrm{healthy})$ of each test image $x$ and its true label. From this, we obtained an AUC of 0.950.

In the second scenario, a reference set of unhealthy examples $D_\mathrm{cancer}$ is available in addition to $D_\mathrm{healthy}$, allowing us to simply classify some $x$ according to which of the reference sets it is closest to. More precisely, we predict a binary unhealthiness label $\hat{y}(x)$ for $x$ via
\begin{equation}
    \hat{y}(x) = \begin{cases}
        0 \text{ if } s(x; D_\mathrm{healthy}) < s(x; D_\mathrm{cancer})\\
        1 \text{ else}\\
    \end{cases}.
\end{equation}
We then aggregate the results over the entire test set via the AUC given the predicted labels and the true labels, resulting in an AUC of 0.989. 

Both of these experiments indicate that the features captured by FRD are highly discriminative for this task, which points to the meaningfulness of the features for medical imaging domains and respective diagnostic applications. Understandably, the task was slightly easier in the second scenario, due to the availability of unhealthy cases for direct comparison.

\section{Properties of FRD}
\label{sec:properties}
In the following sections, we will demonstrate various intrinsic properties of FRD, including its relationship with user (radiologist) preference for image quality (Sec. \ref{sec:userstudy}), computational stability (Sec. \ref{sec:sampleeff}), computational efficiency (Sec. \ref{sec:computetime}), and sensitivity to image corruptions (Sec. \ref{sec:distort}) and adversarial attacks (Sec. \ref{sec:advattack}).

\subsection{Relationship of FRD to User Preference of Image Quality}
\label{sec:userstudy}

In this section, we investigate whether FRD (and other metrics) correlate with human expert-perceived quality of synthetically-generated images, in the context of breast cancer screening. We utilized the experimental design of \citet{garrucho2023high}, where three experienced readers were asked to rate the realism of sets of high-quality synthetic mammography images generated by a CycleGAN \citep{cyclegan} trained on images from a specific acquisition site and anatomical view. This was conducted for six (anatomical view, acquisition site) settings, where the anatomical view is CC or MLO, and the dataset is OPTIMAM \citep{halling2020optimam}, CSAW \citep{csaw}, or BCDR \citep{lopez2012bcdr}. These three readers were two breast radiologists with over 9 and 11 years of experience, respectively, and a surgical oncologist with over 14 years of experience in image-guided breast biopsy. 

\begin{figure}[htbp!]
    \centering
    \includegraphics[width=0.99\linewidth]{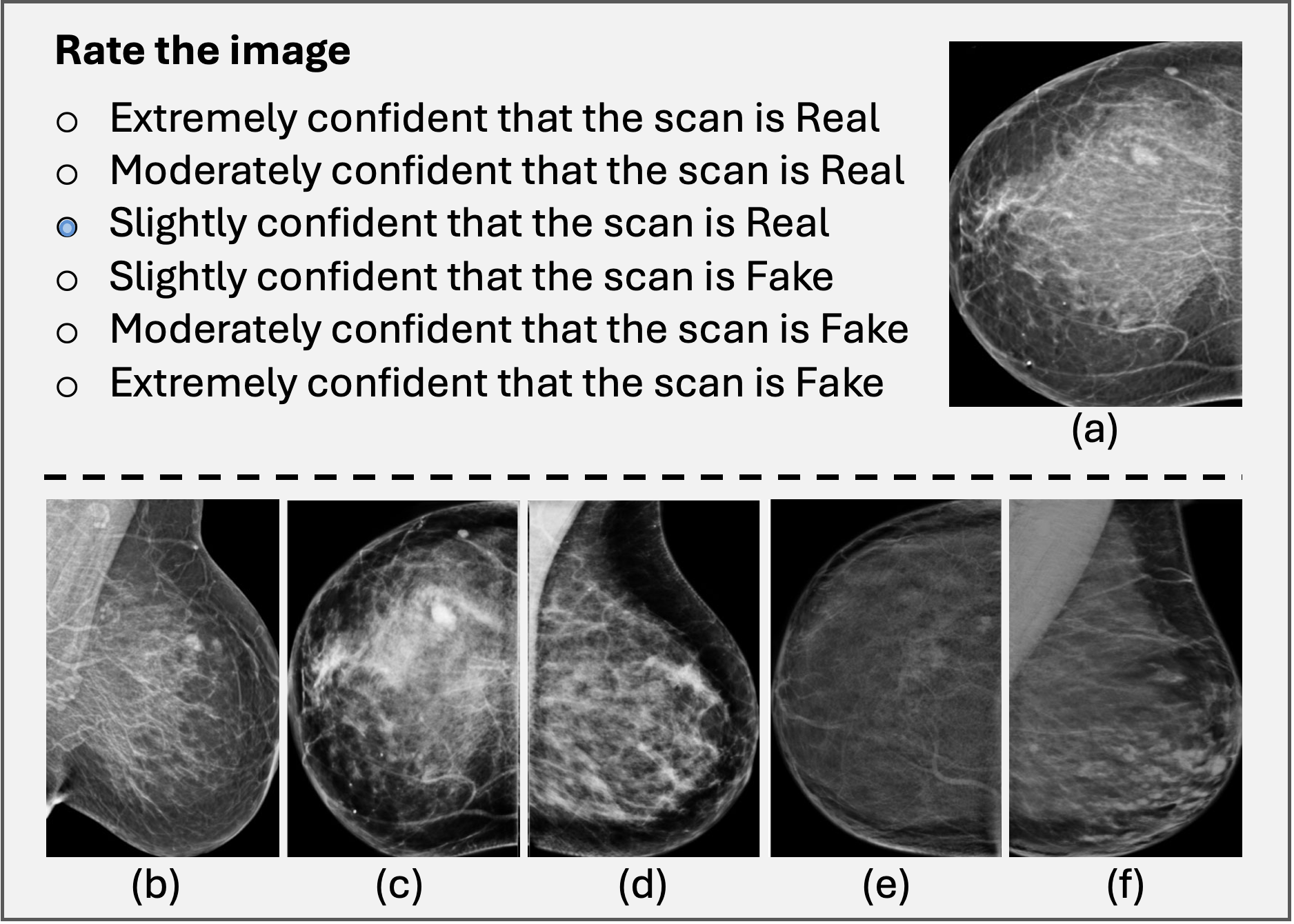}
    \caption{\textbf{Reader Study:} Using an ImageJ plugin, each reader was shown one image at a time, which was randomly sampled from a set of 90 synthetic mammograms. The image set was equally distributed between craniocaudal (CC) and mediolateral oblique (MLO) views (45 each), as well as between the Optimam, CSAW, and BCDR datasets (30 each). Images (a) to (f) are synthetic samples that looked realistic to the readers: (a) BCDR CC, (b) BCDR MLO, (c) Optimam CC, (d) Optimam MLO, (e) CSAW CC, and (f) CSAW MLO. Figure based on \citet{garrucho2023high}.}
    \label{fig:reader_study}
\end{figure}

As illustrated in Fig. \ref{fig:reader_study}, ratings were conducted on a Likert scale, where the user had to rate a given generated image on a scale of 1 (extremely confident in the scan appearing fake) to 6 (extremely confident in it appearing real), where 1, 2, 3, 4, 5, and 6 are respectively mapped to equally-distributed probabilities (final ratings) of 0.05, 0.23, 0.41, 0.59, 0.77, and 0.95 to compute the ROC curve of each reader as in \citet{alyafi2020quality}. In total, the study was completed on 90 synthetic images, containing 15 CC and 15 MLO images from each of the three mammography datasets.

\begin{figure}
\centering
\begin{tabular}{l||ccc}
    \multicolumn{1}{c}{} & \multicolumn{3}{c}{\textbf{Absolute Rating}} \\
    \textbf{Metric} & Pearson $r$ & Spearman $r$ & Kendall $\tau$ \\\toprule
    FRD & -0.65 & -0.71 & -0.6 \\
    FRD$_\text{\textbf{v0}}$ & -0.66 & -0.43 & -0.20\\
    FID & 0.37 & 0.66 & 0.47 \\
    RadFID & 0.91 & 0.94 & 0.87 \\
    \bottomrule
\end{tabular}
\includegraphics[width=0.95\linewidth]{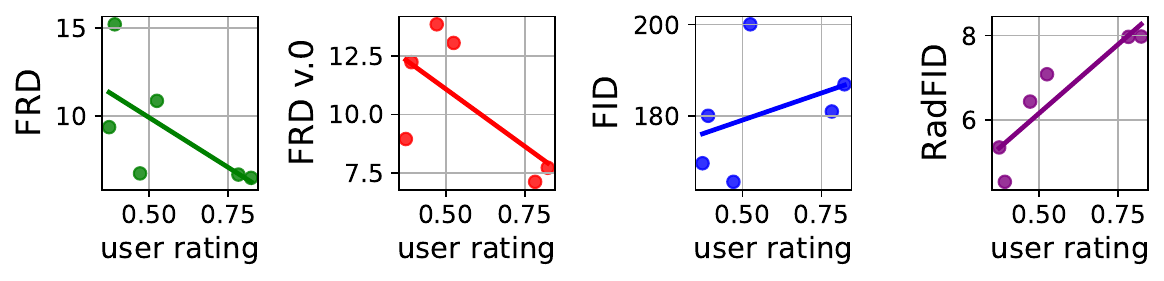}
\caption{\textbf{Top:} Correlation coefficients (linear Pearson $r_P$ and non-linear/rank Spearman $r_S$ and Kendall $\tau$) of different distance metrics with average user (radiologist) preference, for the task of measuring synthetic image quality.  \textbf{Bottom:} Associated plots and linear best fits for this data.}

\label{fig:userstudy}
\end{figure}

Since this study essentially asked readers to visually compare the synthetic images to their known mental reference for how real mammography images generally should appear, we can assess whether FRD and the other perceptual metrics can exhibit the same behavior. To this end, we compute the distance between (a) the set of synthetic images for a given dataset and view and (b) a fixed reference set of 90 real mammography images sampled evenly from all views and datasets\footnote{Note that it is crucial for the reference set to remain constant for these metrics to have fixed scales.}. Shown in Fig. \ref{fig:userstudy}, we analyzed this for each of FRD, FRD$_\text{\textbf{v0}}$, FID, and RadFID, and measured the correlation (Pearson's linear correlation $r$, as well as Spearman's non-linear/rank $r$ and Kendall's $\tau$) between the metric and user rating, where each datapoint corresponds to each possible (view, dataset) combination. User ratings are averaged over all three readers and all 15 synthetic images for the given view and dataset. Additionally, we explore using user ratings \textit{calibrated} by ratings of real data in \ref{app:readerstudycalibrated}.

The desired behavior for a perceptual distance metric is for a metric to negatively correlate with user preference, as higher-quality/more realistic synthetic images---as measured by higher user rating---should correspond to the synthetic images having lower perceptual distance with respect to real images. Shown in Fig. \ref{fig:userstudy}, we see that out of all metrics, on average over all three correlation measures, FRD performs best in this regard, similarly to FRD$_\text{\textbf{v0}}$ in linear correlation ($r\simeq-0.65$) and out-performing noticeably in non-linear correlation. 


On the other hand, FID and RadFID actually \textit{anti-correlate} with user perception of quality of synthetic images, particularly RadFID; FID is surprisingly less worse in this regard (aligning with recent similar findings in \citet{woodland2024_fid_med}), despite RadFID utilizing pretrained domain-specific medical image features, while FID uses natural image features. This surprising result for RadFID could be due to its underlying Inception encoder being trained to focus on very specific patterns to detect (often localized) disease in the pretraining images and tasks of RadImageNet \citep{radimagenet}. While such local patterns may have high influence on RadFID, they may have relatively small influence on the overall image appearance and therefore the readers' ratings of overall image quality, hence the seeming ``mismatch'' between RadFID and the readers' ratings. On the contrary, FRD captures many generic, global image features (that didn't rely on some training set) that will likely end up overlapping with reader perception in some aspect, which may be why FRD (negatively) correlates more predictably with reader preference. Overall, this provides further evidence that FID and RadFID should be used with caution for medical images, and that FRD provides a noticeable alternative that correlates substantially better with the perceptual preference of experienced readers.

\subsection{Sample Efficiency and Stability}
\label{sec:sampleeff}
The stability of perceptual metrics at small sample sizes is key for medical image datasets due to them being typically smaller (\eg, \( N \approx 10^2 \)--\( 10^4 \)) than natural image datasets (\eg, ImageNet with \( N \approx 10^6 \)). FID generally requires \( N \approx 10^4 \)--\( 10^5 \) samples for stability \citep{fid, jayasumana2023rethinking}, which can be prohibitive in this setting. We will now evaluate FRD, FRD$_\text{\textbf{v0}}$, RadFID, and FID across varying sample sizes \( N \) to test for this stability.

\begin{figure}[htbp!]
    \centering
    \includegraphics[width=0.99\linewidth]{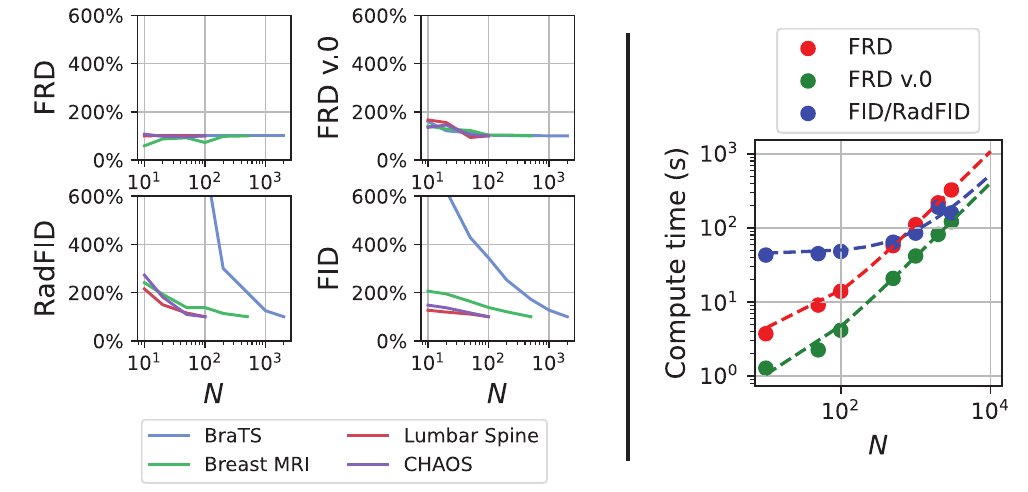}
    \caption{\textbf{Left:} Sensitivity of FRD, FRD$_\text{\textbf{v0}}$, RadFID and FID to sample size $N$. Metric values (vert. axes) are relative to their highest-$N$ result. \textbf{Right:} Computation time for the metrics w.r.t $N$, with linear best fits plotted in dashed lines.}
    \label{fig:combined_sampleeff}
\end{figure}

We test this under the case of image translation evaluation for the main datasets (Table \ref{tab:datasets}): CycleGAN for breast MRI, GcGAN for brain MRI, CUT for lumbar spine, and MUNIT for CHAOS, respectively. Shown in Fig. \ref{fig:combined_sampleeff} left, FRD remains stable even for very small $N$ (down to \( N = 10 \)), while RadFID and FID---as well as FRD$_\text{\textbf{v0}}$, to a lesser extent---diverge as $N$ grows small across all datasets, indicating that these FID-based metrics are not suitable for comparing small medical datasets of different sizes. 

The stability of FRD at small \( N \), especially compared to that of FID and RadFID, is likely due to relatively few features dominating its computation (see \eg, Fig. \ref{fig:radiomic_interp}(a)), meaning the effective dimensionality \( \tilde{m} \) is much smaller than the full \( m \approx 500 \). Thus, the \Frechet{} distance in FRD behaves as though it operates in a lower-dimensional space, enhancing stability even with limited samples.

More concretely, consider that computing the \Frechet{} distance (Eq. \ref{eq:frechet}) as is necessary for FRD and FID requires estimating the covariances matrices $\Sigma_1, \Sigma_2\in\R^{m\times m}$ of the two data distributions in the $m$-dimensional feature space (\( m \approx 500 \) for FRD while $m\approx 2000$ for FID). For $N\ll m$, the estimation of these matrices can become very unstable, \textit{assuming that} most of the $m$ feature dimensions contribute similarly to the variability of the data distributions, \ie, $\tilde{m} \simeq m$. 

The computational stability results shown in Fig. \ref{fig:combined_sampleeff} indicate $\tilde{m}$ to be typically higher for FID/RadFID features compared to FRD features, as the former are noticeably less stable for low $N$. As FRD features appear to have relatively low $\tilde{m}$, the true covariance matrices $\Sigma$ in this feature space are approximately low rank, so that the estimation is stable for small $N$ because they can be mostly represented by the first $\tilde{m}$ terms of their eigendecompositions, \eg, $\mathrm{tr}(\Sigma)\simeq \sum_{i=1}^{\tilde{m}}\lambda_i$ given eigenvalues $\lambda_i$ of $\Sigma$.

\subsection{Computation Time}
\label{sec:computetime}
We next compare the computation time of FRD to FID/RadFID (as well as FRD$_\text{\textbf{v0}}$) across sample sizes \( N \), using data parallelism (\texttt{num\_workers=8}) on UNSB-translated BraTS test images. As shown in Fig. \ref{fig:combined_sampleeff} right, FRD is faster than FID/RadFID for small-to-moderate sample sizes (\( N \lesssim 500 \)). For larger \( N \), the computation time of FRD grows slightly faster, but both metrics remain efficient across different sample sizes (the computation time scales linearly with $N$ asymptotically), with FRD and FRD$_\text{\textbf{v0}}$ particularly advantageous for small \( N \). While FRD is slightly slower than FRD$_\text{\textbf{v0}}$---owing to the addition of various image filters prior to radiomic computation (Sec. \ref{sec:radiomicdist})---the improved performance in essentially all tested applications (Sec. \ref{sec:experiments}) make it well worth it. 

\subsection{Sensitivity to Image Corruptions}
\label{sec:distort}
In this section, we analyze the sensitivity of FRD to image corruptions that may affect downstream task performance, compared to prior common metrics. Given the importance of downstream task performance metrics for medical image translation models, as well as the typical increased sensitivity of medical image models to image corruptions compared to natural image models \citep{konz2024effect}, we study if a corruption to a (translated) image that noticeably affects downstream task performance on that image is also captured by the perceptual metrics. For further results, see \ref{app:MRIdistort} where we show that FRD and FRD are sensitive to various realistic image corruptions in MRI.

Consider some image transformation/corruption $T:\R^n\rightarrow\R^n$.
We model a preferable perceptual metric $d$ as approximately following the inverse proportionality
\begin{equation}
      \frac{d(\Dt, T(\Dst))}{d(\Dt, \Dst)} \underset{\sim}{\propto} \left(\frac{\mathrm{Perf}(T(\Dst))}{\mathrm{Perf}(\Dst)}\right)^{-1},
\end{equation}
evaluated on the test set's target domain images $\Dt$ and translated source-to-target images $\Dst$. In other words, if the perceptual distance increases by some positive multiplicative factor $K$ (implying the corruption made the translated images more distant from the target domain), we would expect the performance to go worse by $1/K$, up to a constant of proportionality---the sensitivity of the two metrics to image corruptions should match.

\begin{figure}[thpb]
    \centering
    \includegraphics[width=0.95\linewidth]{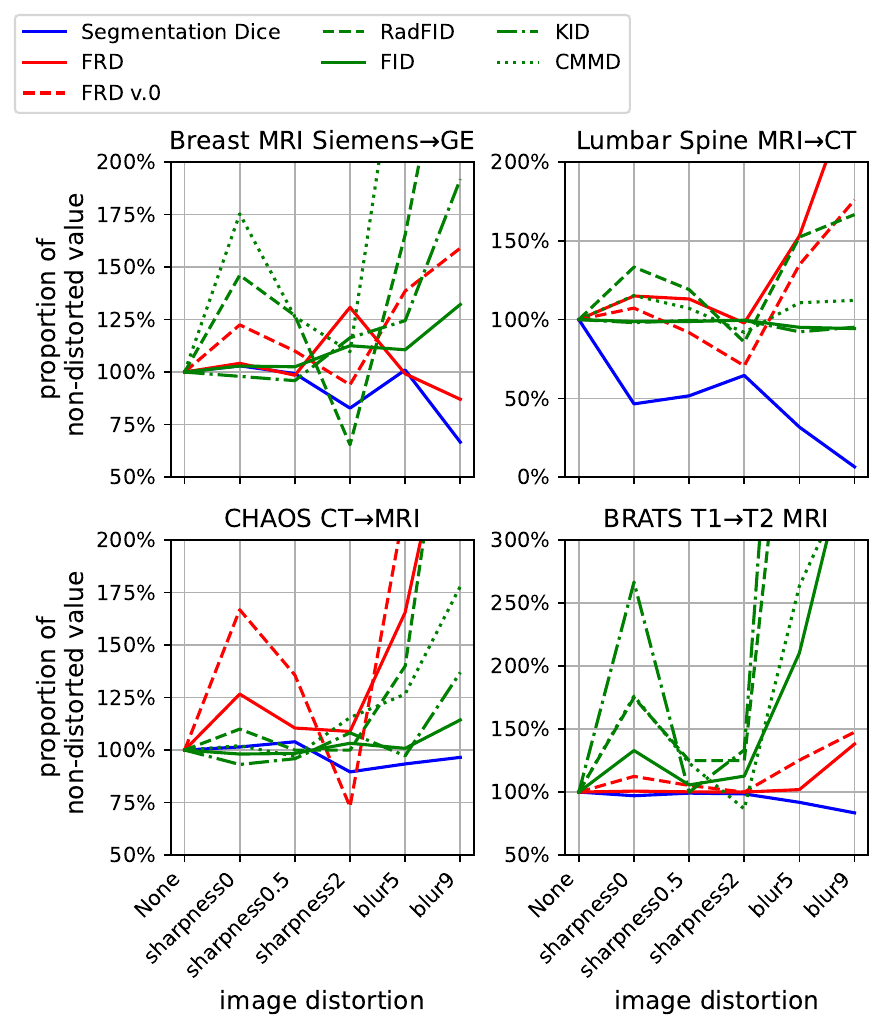}
    \caption{Sensitivity of FRD, FRD$_\text{\textbf{v0}}$ (\textcolor{red}{\textbf{red lines}}), RadFID, FID, KID, and CMMD (\textcolor{ForestGreen}{\textbf{green lines}}) to corruptions which affect downstream task performance (\textcolor{blue}{\textbf{blue line}}) on translated images. Metric values (vertical axis) are relative to their un-distorted result (``None'').}
    \label{fig:distortions}
\end{figure}

We tested this on all datasets for downstream segmentation tasks that were sensitive to such corruptions (fibroglandular tissue (FGT) for breast MRI, tumor for brain MRI, bone for lumbar spine, and liver for CHAOS (Table \ref{tab:results_tasks})), for various translation models (CycleGAN, GcGAN, CUT, and MUNIT, respectively). We evaluate simple image corruptions $T$ as Gaussian blurs with positive integer kernel $k$ (labeled as \texttt{blur$k$}) or sharpness adjustments of non-negative factor $\gamma$ (\texttt{sharpness$\gamma$}), from TorchVision \citep{torchvision2016}. We show the results in Fig. \ref{fig:distortions}, plotting ${d(\Dt, T(\Dst))}/{d(\Dt, \Dst)}$ and ${\mathrm{Perf}(T(\Dst))}/{\mathrm{Perf}(\Dst)}$ for each corruption for all proposed and prior perceptual metrics $d$.

For breast MRI, we see that the corruption sensitivity of FID and FRD fairly well match the performance sensitivity, with the other perceptual metrics less so (for example, FRD and RadFID can be over-sensitive to corruptions which barely affect performance). For lumbar spine, FRD, FRD$_\text{\textbf{v0}}$ and RadFID follow the performance sensitivity well, while other perceptual metrics are generally not as sensitive. For BraTS, performance is typically not sensitive to corruptions, which FRD and FRD$_\text{\textbf{v0}}$ follow; other perceptual metrics are generally oversensitive, but all still increase when performance decreases due to blurring. CHAOS is an interesting case, where all perceptual metrics except for FID are overly sensitive to corruptions despite only small changes to performance, likely due in part to challenges of training the downstream task model on such a small and challenging dataset (Sec. \ref{sec:datasets}). Overall, the sensitivity of FRD aligns best with performance sensitivity. Other perceptual metrics are not as consistent over all datasets, which aligns with the results of Sec. \ref{sec:exp:downstreamcorr}.

\subsection{Sensitivity to Targeted Adversarial Attacks}
\label{sec:advattack}

Another type of image corruption that is important to consider is adversarial examples, a type of adversarial attack on some trained downstream task model $f$ where an input image $x$ is modified in a targeted manner to drastically change the model's prediction $f(x)$ to be incorrect \citep{advexamples}, while constraining the modifications to $x$ to be as subtle as possible. This scenario is important to consider due to the safety-critical nature of medical image diagnosis applications. Here, we will consider the case of attacking a binary classification neural network to judge whether FRD (and other distance metrics) can detect when images have been attacked.

We use the FGSM method \citep{advexamples} to attack some input image $x$; FGSM uses the gradient of the network's prediction loss to modify the input image $x$ with true binary domain label $y$ to some $\tilde{x}$ in order for $f$ to misclassify $x$, as
\begin{equation}
    \tilde{x} = \mathrm{FGSM}(x,y) := x + \epsilon \mathrm{sign}\left(\frac{\partial L(f(x), y)}{\partial x}\right),
\end{equation}
where $L$ is the (binary cross-entropy) loss between the model's prediction $f(x)$ and the true label $y$.
The use of the real-valued parameter $\epsilon>0$ and $\mathrm{sign}$ function constrains the attacked image to be imperceptible with a tolerance of $\epsilon$, \ie, $||\tilde{x}-x||_\infty < \epsilon$. 

Given some test dataset $D$ of images $x$ and corresponding labels $y$, we will attack each image in $D$ to obtain the attacked set $\tilde{D}_\epsilon := \{ \tilde{x} = \mathrm{FGSM}(x,y) : (x,y) \sim D \}$, using various $\epsilon$ (note that $\epsilon=0$ is the baseline case of non-attacked images, \ie, $\tilde{D}_0 = D$). We will then determine if various distance metrics $d$, including FRD, can detect the attacked images compared to a separate reference set of clean images (the training set for $f$) $D_\mathrm{ref}$, via $d(D_\mathrm{ref}, \tilde{D}_\epsilon)$. We will measure two desiderata: \textbf{(a)} if the metrics can differentiate between clean and attacked images---\ie, if $d(D_\mathrm{ref}, D) < d(D_\mathrm{ref}, \tilde{D}_\epsilon)$ for various $\epsilon$---and \textbf{(b)} if the distances $d(D_\mathrm{ref}, \tilde{D}_\epsilon)$ increase as the attack becomes more severe ($\epsilon$ grows larger).

We use a subset of the experimental setup of \citet{konz2024effect} (see that paper for more details), performing attacks on binary classification models for seven medical image datasets. These are (1) brain MRI glioma detection (\textbf{BraTS}, \citet{brats}); (2) breast MRI cancer detection (\textbf{DBC}, \citet{saha2018machinedukedbc}); (3) prostate MRI cancer risk scoring (\textbf{Prostate MRI}, \citet{sonn2013prostate}); (4) brain CT hemorrhage detection (\textbf{RSNA-IH-CT}, \citet{flanders2020rsnaihct}); (5) chest X-ray pleural effusion detection (\textbf{CheXpert}, \citet{irvin2019chexpert}); (6) musculoskeletal X-ray abnormality detection (\textbf{MURA}, \citet{rajpurkar2017mura}); and (7) knee X-ray osteoarthritis detection (\textbf{OAI}, \citet{Tiulpin2018}). We evaluate $f$ as being either a ResNet-18 \citep{he2016resnet} or a VGG-13 \citep{VGGs}, and train them with training sets ($D_\mathrm{ref}$) of size $1750$, before evaluating and attacking them on class-balanced test sets of size $750$, via the dataset creation/sampling procedures described in \citet{konz2024effect}. As shown in full in \ref{app:egattacks}, these attacks are typically quite successful, even for very small values of $\epsilon$, despite them being practically undetectable visually. 

\begin{figure*}[thpb]
    \centering
    \includegraphics[width=0.95\linewidth]{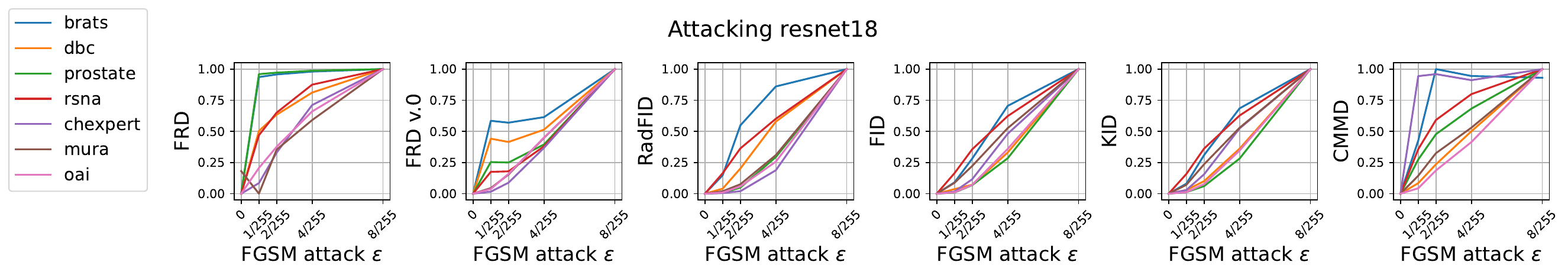}
    \includegraphics[width=0.95\linewidth]{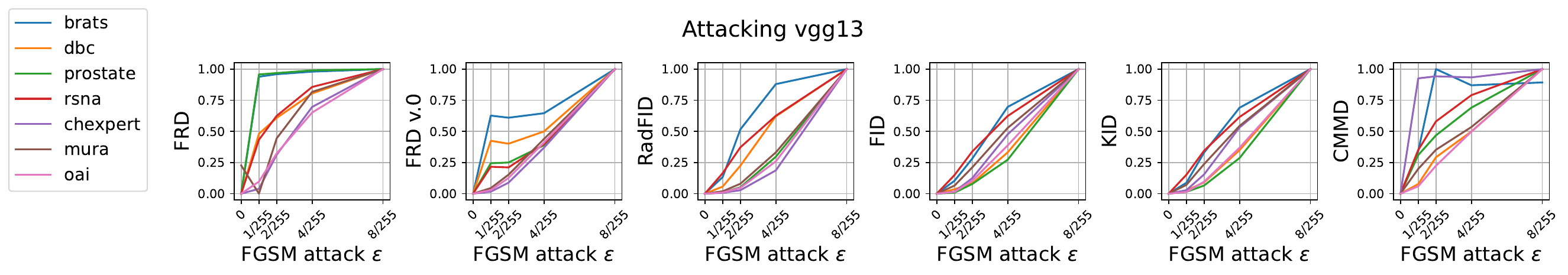}
    \caption{Using different distance metrics (vertical axes) to detect FGSM adversarial attacks of various strength (horizontal axes), for ResNet-18 (top) and VGG-13 (bottom) classification models. Note that distance metrics are scaled to $[0,1]$ over their plotted range of values for the sake of visualization.}
    \label{fig:advatk_dist}
\end{figure*}

We show the results of these experiments in Fig. \ref{fig:advatk_dist}, where we show how $d(D_\mathrm{ref}, \tilde{D}_\epsilon)$ changes with respect to attack strength $\epsilon$ (including $\epsilon=0$ for the unattacked case of $\tilde{D}_\epsilon = D$), for various distance metrics $d$, on all datasets and models. We see that FRD clearly differentiates attacked images ($\epsilon>0$) from non-attacked images ($\epsilon=0$) in almost all cases (desiderata (a)), and that it typically increases with higher attack strength/$\epsilon$ (desiderata (b)). Moreoever, FRD is typically more sensitive to the most subtle attacks ($\epsilon=1/255$) compared to FRD and almost all other metrics, possibly due to the inclusion of frequency/wavelet features. These results show that FRD can detect adversarial attacks and their severity, despite the fact that these attacks are practically imperceptible.

\section{FRD for Interpretability}
\label{sec:interp}
In this section, we will demonstrate how FRD aids in interpreting differences between large sets of medical images, \ie, understanding the main features that differ between the two sets. The example we will study is interpreting the effects of image-to-image translation models, but this formalism could be applied to any two distributions of images.

At the single-image level, an input image $x_s$ and output translated image $x_{s\rightarrow t}$ can be converted to feature representations (radiomic or learned) $h_s:=f(x_s)$ and $h_{s\rightarrow t}:=f(x_{s\rightarrow t})$, and we can interpret the feature change vector $\Delta h:= h_{s\rightarrow t} - h_s$ and it's \textit{absolute change} counterpart $|\Delta h|$ defined element-wise by $|\Delta h|^i := |h^i_{s\rightarrow t} - h^i_s|$. At the image distribution level, we can define $\Delta h := \mu_{s\rightarrow t} - \mu_s$ (and similarly $|\Delta h|$ via $|\Delta h|^i := |\mu^i_{s\rightarrow t} - \mu^i_s|$), where $\mu_s, \mu_{s\rightarrow t}\in \R^m$ are the mean vectors of the input and output feature distributions, respectively. In this case, we also define the distributions of values for \textit{individual features} as $F_s^i:=\{h_s^i : h_s\in F_s\}$ and $F_{s\rightarrow t}^i:=\{h_{s\rightarrow t}^i : h_{s\rightarrow t}\in F_{s\rightarrow t}\}$.

In either case, $\Delta h$ is simply the linear direction vector in feature space between the input and output distributions, analogous with other interpretability works that utilize the linear representation hypothesis \citep{parklinear,tcav,alain2017understanding,konz2023attributing}. We will next discuss the options and challenges for interpreting $\Delta h$, for either learned features or fixed (radiomic) features.

\begin{figure}[thpb]
    \centering
    \includegraphics[width=0.99\linewidth]{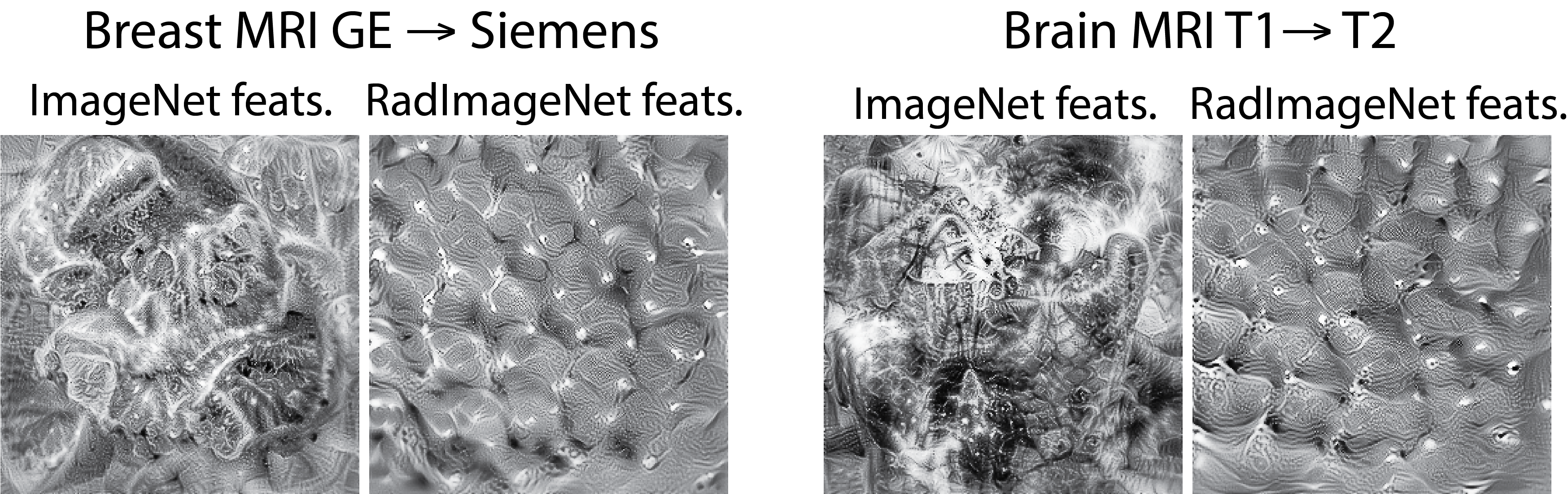}
    \caption{Attempts at medical image translation interpretability via learned feature inversion.}
    \label{fig:translation_inversion}
\end{figure}

\paragraph{Attempting Interpretability with Learned vs. Radiomic Features}
A common method for interpreting directions $v$ in a deep encoder’s feature space, such as $\Delta h$, is \textit{feature inversion} \citep{olah2017feature, featureinvert}, which uses gradient-based optimization to find an input image \( x_v \) that aligns with \( v \) in feature space, \ie,
\begin{equation}
x_v = \mathrm{argmax}_{x} \mathrm{cossim}(v, f(x)). 
\end{equation}
However, we found that doing so using either ImageNet or RadImageNet features resulted in abstract visualizations that lack clear, quantitative insights useful for clinical interpretation (Fig. \ref{fig:translation_inversion}; see \ref{app:interp_featureinversion} for details).

Alternatively, the \textit{individual} features of \( \Delta h \) could be examined statistically with questions like ``\textit{which features changed the most?}'' or ``\textit{did only a few features account for most of the cumulative change?}''. However, concretely interpreting individual \textit{learned} features remains challenging due to the qualitative nature of feature inversion, so we face the same problem.

Thankfully, the clear definitions of radiomic features (Sec. \ref{sec:radiomicdist}) allow for clear, quantitative answers to feature interpretability questions, beyond what is possible for learned feature techniques like feature inversion. Here we will exemplify this by interpreting a CUT model trained for lumbar translation, with the following questions.

\begin{figure}[thpb]
    \centering
    \includegraphics[width=0.99\linewidth]{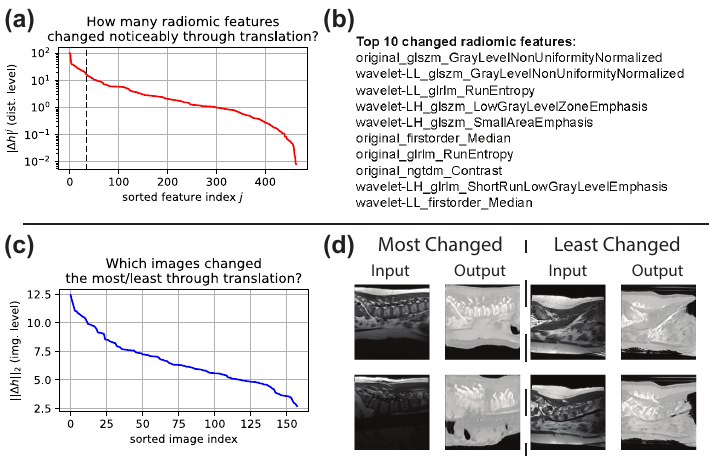}
    \caption{\textbf{Translation interpretability using radiomic features.}}
    \label{fig:radiomic_interp}
\end{figure}

\begin{enumerate}
\item \textbf{Which features changed the most?} Sorting features by their values in the \( |\Delta h| \) between input and output image distributions (Fig. \ref{fig:radiomic_interp}(a)) identifies those with the highest change, primarily textural/gray-level matrix features, reflecting appearance shifts from MRI to CT (Fig. \ref{fig:radiomic_interp}(b)).
\item \textbf{Did only a few features change significantly?} Yes---50\% of cumulative feature changes (measured by $|\Delta h|$) are covered by only 37 out of 500 features, indicating a light-tailed distribution (Fig. \ref{fig:radiomic_interp}(a)).
\item \textbf{Which images changed the most or least?} Sorting input/output image pairs $(x_s, x_{s\rightarrow t})$ by their absolute feature change $||h_{s\rightarrow t} - h_s||_2=||\Delta h||_2$ (Fig. \ref{fig:radiomic_interp} (c)) shows that the most-changed images have distinct anatomical differences, while the least-changed images mainly differ in texture and intensity (Fig. \ref{fig:radiomic_interp}(d)).
\end{enumerate}

This interpretability methodology could also help compare translation models on the same dataset and assess model effects on the images, or analyze the domain shift between two datasets.

\section{Discussion and Conclusions}
\label{sec:discussion}

Overall, our results with FRD show the value of using interpretable, medical image-specialized feature spaces like radiomic features for comparing unpaired medical image distributions. We showed that FRD consistently and substantially 
improves over FRD, FID, RadFID, as well as other prior common metrics for unpaired medical image comparison on 
all evaluated tasks. These include stronger alignment with downstream task metrics and anatomical consistency (Sec. \ref{sec:exp:downstreamcorr}) for the task of evaluating image-to-image translation, OOD detection ability (Sec. \ref{sec:ood}), alignment with radiologist preferences of image quality (Sec. \ref{sec:userstudy}), computational stability and efficiency, especially for small sample sizes (Secs. \ref{sec:sampleeff} and \ref{sec:computetime}), sensitivity to performance-affecting image corruptions (Sec. \ref{sec:distort}) and adversarial image attacks (Sec. \ref{sec:advattack}), as well as, notably, enhanced interpretability for clinical use (Sec. \ref{sec:interp}).

\paragraph{Medical Image Similarity from the Perspective of Radiologists}
It is also important to consider the task of measuring the similarity of medical images from the perspective of radiologists. To explore this, we performed a semi-structured expert interview of two experienced radiologists (authors L.G. and J.L.). According to them, distinguishing between images from clearly different acquisition types, such as T1- vs. T2-weighted brain MRI or between modalities like abdominal/lumbar spine MRI and CT scans (Fig. \ref{fig:datasets}), is a straightforward task. In contrast, differentiating between more subtly distinct domains, such as breast MRI scans acquired using the same sequence type but different scanner manufacturers or settings (Fig. \ref{fig:datasets}), is considerably more challenging. Nonetheless, identifying such differences is crucial, as even subtle domain shifts can significantly impact the performance of downstream tasks (Table \ref{tab:results_tasks}, bottom two rows), despite the almost imperceptible visual differences between domains. 

Notably, FRD is able to detect these subtle domain differences with high consistency (see the OOD detection results of Table \ref{tab:ood}, first column). If a radiologist was to use such a downstream task model in clinical practice, FRD could help inform whether the model is well-calibrated for some new image at hand by detecting if the new image is OOD of the model’s training set. More generally, this could be used to detect any gradual distribution shift of newly acquired data over time compared to the model’s original training set, indicating if the model may require fine-tuning on the new data.

\paragraph{Recommendations for Best Medical Image Translation Models} 
A further contribution of our study comprises a comprehensive set of experiments to compare source-to-target domain image translation methods. To this end, our empirical findings suggest notable general recommendations for medical image translation models. For severe domain shifts (\eg, lumbar spine and CHAOS), CUT performed best in both downstream tasks and perceptual metrics of FRD and RadFID, likely due to its contrastive learning approach that preserves image structure. For more subtle shifts (breast MRI and brain MRI), GcGAN performs well, with UNSB and CycleGAN also effective for breast MRI. As such, our findings recommend incorporating contrastive learning approaches such as CUT for high domain shifts, while adversarial learning may suffice for moderate shifts, as demonstrated by GcGAN.

\paragraph{Limitations} 
Our work covered a range of diagnostic tasks and imaging domains, but is still limited to radiology. Other modalities 
could be explored, though radiomic features may need adjustment or (partial) replacement by other types of modality-specific imaging biomarkers.
In this regard, our work demonstrates the vast potential of using imaging biomarker variability as a dataset comparison technique across medical domains. 

For instance, in histopathology \citep{gurcan2009histopathological, gupta2019emergence,holscher2023next} and cytopathology \citep{rodenacker2003feature}, quantitative pathomics features can be extracted, characterizing tissue architecture, cellular morphology, cell density, texture, intensity, and spatial relationships. Therefore, measuring the variability between such pathomics feature distributions across sets of cytology or histology images or image patches can provide insights into out-of-distribution detection, tissue characterization, disease localization, and generative model evaluation in this domain.
This motivates future work to define and empirically assess variants of a respective ``Fréchet Pathomic Distance''. On another note, in further medical imaging domains, such as dermatology, radiomic features have already shown utility in disease characterization \citep{attallah2021intelligent,wang2024radiomic}, suggesting such features and, thus, FRD can be directly applicable for comparing distributions in such domains. Therefore, there is much potential for future work in quantifying the descriptive power of radiomics across different medical fields to identify where additional biomarkers are needed and where radiomics features suffice, followed by experiments evaluating FRD in these contexts.

Finally, our interpretability contributions are nascent, and further work is needed to extract more qualitative, yet concrete, insights. 
Additionally, the absolute values of FRD are only meaningful in relative comparisons between models on the same dataset (\ie, measured using the same reference dataset $D_1$), similar to FID and other metrics. 

\paragraph{Future Work and Expansions}
There are many potential further applications of FRD with one example being the evaluation of multimodal models, such as vision-language 
models, \eg, CLIP-type \citep{clip} models or text-to-image generative models, \eg, Stable Diffusion-type \citep{stablediffusion} models. FRD may also hold potential for evaluating video generative models (e.g., in ultrasound imaging \citep{chen2024ultrasound, reynaud2025echoflow}), prompting further research into optimal strategies for feature handling—such as whether to (a) extract aggregated features from video sequences or (b) aggregate features extracted from individual frames. Additionally, applying FRD to sets of video frames could support identification of frames of interest and detection of transitions where the informational content of the video changes.

Future work can consider certain modifications and expansions to FRD. For instance, radiomic features can be computed exclusively within a mask or region of interest (ROI), and, as another alternative, subsequently even further extended to a weighted combination of local (within-mask) and global (whole-image) features. This would enable additional applications, such as evaluating the quality or variability of image annotations by computing FRD between radiomic features extracted from a “gold standard” mask and those from candidate annotations. Such an approach would not only quantify the impact of inter- and intra-observer variability on radiomic biomarkers but also offer a complementary way to assess segmentation and object detection model performances. This assessment can be complementary alongside traditional annotation similarity metrics such as the Dice Score, Intersection over Union and Hausdorff Distance, which can be overly influenced by annotation size, spatial extent, and outliers \citep{reinke2024understanding}.
On another note, the effect of weighting certain types of features (\eg, image-space, frequency-space/wavelet, etc.) over others within FRD could be modified, which may be advantageous for certain applications.


\section*{Acknowledgments}

Research reported in this publication was supported by the National Institute Of Biomedical Imaging And Bioengineering of the National Institutes of Health under Award Number R01EB031575.
The content is solely the responsibility of the authors and does not necessarily represent the official views of the National Institutes of Health. 
This research received funding from the European Union’s Horizon Europe and Horizon 2020 research and innovation programme under grant agreement No 101057699 (RadioVal) and No 952103 (EuCanImage), respectively. It was further partially supported by the project FUTURE-ES (PID2021-126724OB-I00) and AIMED (PID2023-146786OB-I00) from the Ministry of Science, Innovation and Universities of Spain. Richard Osuala acknowledges a research stay grant from the Helmholtz Information and Data Science Academy (HIDA). Daniel M. Lang and Julia A. Schnabel received funding from HELMHOLTZ IMAGING, a platform of the Helmholtz Information \& Data Science Incubator.

\section*{Author Information}
\label{sec:authoraffil}
The affiliations for the author list on the title page are as follows:\\
\small{$^1$ Department of Electrical and Computer Engineering, Duke University, USA}\\
\small{$^2$ Dept. de Matemàtiques i Informàtica, Universitat de Barcelona, Spain}\\
\small{$^3$ Institute of Machine Learning in Biomedical Imaging, Helmholtz Center Munich, Germany}\\
\small{$^4$ School of Computation, Information and Technology, TU Munich, Germany}\\
\small{$^5$ Department of Radiology, Duke University, USA}\\
\small{$^6$ Department of Biostatistics \& Bioinformatics, Duke University, USA}\\
\small{$^7$ Department of Computer Science, Duke University, USA}\\
\small{$^8$ Department of Biomedical Engineering, Yale University, USA}\\
\small{$^9$ Department of Radiology, Weill Cornell Medical College, USA}\\
\small{$^{10}$ Department of Radiology \& Biomedical Imaging, Yale University, USA}\\
\small{$^{11}$ Department of Electrical Engineering, Yale University, USA}\\
\small{$^{12}$ School of Biomedical Engineering \& Imaging Sciences, King's College London, UK}\\
\small{$^{13}$ Computer Vision Center, Universitat Autònoma de Barcelona, Spain}\\
\small{$^{14}$ ICREA, Barcelona, Spain }\\

{
    \small
    \bibliographystyle{ieeenat_fullname}
    \bibliography{main}
}

\newpage

\clearpage
\section*{Supplementary Material}
\setcounter{page}{1}
\appendix

\section{Dataset and Task Labeling Details}
\label{app:datasets}

\subsection{Breast MRI}
For breast MRI we use the 2D slices of the pre-contrast scan volumes from the Duke Breast Cancer dataset \citep{saha2018machine}, using the same train/validation/test splits (by patient) and preprocessing of \citep{segdiff} from the 100 patient volumes with FGT and breast segmentation annotations (see the following paragraph). This results in train/validation/test splits with {source, target} domain sub-splits of size $\{4096, 7900\}/\{432,1978\}/\{688,1890\}$ images.
\paragraph{FGT and breast segmentation} FGT (fibroglandular/dense tissue) and breast segmentation masks for this dataset are provided from \citep{lew2024publicly}.
\paragraph{Cancer classification/slice-level detection} For the cancer classification task, we follow the same convention of \citep{konz2024reverse}, and label slice images as cancer-positive if they contain any tumor bounding box annotation, and negative if they are at least $5$ slices away from any positive slices (ignoring the intermediate ambiguous slices). We then train a basic ResNet-18 \citep{he2016resnet} (modified for 1-channel input images) as our binary cancer classification model, on the positive and negative slices from the training set's GE scans. The model's evaluation datasets are otherwise unchanged from the other downstream tasks (besides the labels used for the images).

\subsection{Brain MRI}
For brain MRI, we utilized the multi-modal brain tumor dataset from the BraTS 2018 challenge \citep{brats}. Since the BraTS's own validation set doesn't have masks available, we began by extracting the original shared training set and dividing the patients into training, validation, and test sets with a ratio of 0.7:0.15:0.15 for this paper. Next, we focused on the T1 and T2 sequence volumes along with their corresponding masks. Each slice of the image volume was normalized and saved as 2D PNG files to construct our 2D dataset. Note that because by default, each patient has both T1 and T2 scans, we used randomly sampling to construct the T1 and T2 subsets of the train, validation, and test sets such that there is no overlap between patients for the T1 and T2 sets (for example, the T1 and T2 test set images).

\paragraph{Tumor segmentation and detection} The original mask contains multiple classes of segmentation, including: Background (Label 0), Enhancing Tumor (Label 4), Tumor Core (Label 1), Whole Tumor (Label 2), Peritumoral Edema (Label 3). We conducted a binary tumor/not-tumor segmentation by combing all pixels with label larger than 0. For tumor detection, the tumor bounding box is generated by the smallest box that covers the entire tumor region. For those cases without tumor shown in that slice, we excluded them during model training/validation/testing.

\paragraph{Cancer classification/slice-level detection} We also further modify the task into a binary tumor classification task: whether this slice contains tumor or not. For those slices that are near the boundary of the tumor, specifically the 5 slices before and after the tumor presence (switching between positive and negative in each volume), we excluded them from the classification as they are considered ambiguous slices. 

\subsection{Lumbar Spine}
The CT lumbar spine dataset is obtained from TotalSegmentator \citep{Wasserthal_2023}, and the T1 MRI data is private (to be revealed upon paper acceptance). We split the 2D source and target data in train/val/test as $\{495, 1466\}/\{175, 409\}/\{158, 458\}$.

\paragraph{Bone segmentation} We perform binary classification on each pixel to determine whether it includes bone or not. The ground truth masks for MRI are reviewed by experts, while the CT masks are sourced from \citep{Wasserthal_2023}.

\subsection{CHAOS}
We extract 2D CT and T1 in-phase MRI slices from the CHAOS dataset \citep{kavur2021chaos}. For each domain, we randomly split the data by patient in the ratio of 10:5:5, resulting in the 2D slices for the source and target domains being divided into train/val/test as $\{1488, 322\}/\{926, 182\}/\{460, 182\}$.

\paragraph{Liver segmentation} Liver masks for both modalities are provided by \citep{kavur2021chaos}.

\paragraph{Liver classification} We assign positive labels to slices which contain the liver and negative labels to those that do not.

\subsection{Single Breast Cancer Prediction Experiments}
\label{app:frdpredictordata}

For the experiments of Section \ref{sec:FRDpredictor}, we used the MAMA-MIA \citep{mamamia} breast DCE-MRI dataset. For the train and test sets of MAMA-MIA, each 2D slice image from each axial MRI volume was split in half down the middle to result in two images of single breasts. A given single breast image was then labeled as either healthy or cancerous if its accompanying lesion segmentation (from \citet{mamamia}) was non-zero within the image. Applying this procedure to MAMA-MIA's test set resulted in our experiment's test set of both healthy and cancerous single breast images, and applying it to MAMA-MIA's train set resulted in our reference sets of healthy and cancerous images, $D_\mathrm{healthy}$ and $D_\mathrm{cancer}$, respectively.

\section{Model Training/Architectural Details}
\label{app:training}

In this section we describe the training details of all networks in the paper. All experiments were completed on four 48GB NVIDIA A6000 GPUs.

\subsection{Translation Models}
\label{app:training:trans}

All six translation models (CycleGAN \citep{cyclegan}, MUNIT \citep{munit}, CUT \citep{cut}, GcGAN \citep{gcgan}, MaskGAN \citep{maskgan}, and UNSB \citep{unsb}) were trained with their default settings (besides being modified to input and output $1$-channel images), except for a few exceptions to be described shortly; these settings are shown in Table \ref{tab:app:transmodelsdetails}. 

\begin{table*}[htbp]
\centering
\begin{tabular}{l|cc}
\textbf{Model} & \textbf{Training time} & \textbf{Batch size} \\
\midrule
CycleGAN \citep{cyclegan} & 200 epochs & 4 \\
MUNIT \citep{munit} & 1M iters. & 1 \\
CUT \citep{cut} & 200 epochs & 1 \\
GcGAN \citep{gcgan} & 200 epochs & 32 \\
MaskGAN \citep{maskgan} & 200 epochs & 4 \\
UNSB \citep{unsb} & 200 epochs & 1  \\

\bottomrule
\end{tabular}
\caption{Translation model training details.}
\label{tab:app:transmodelsdetails}
\end{table*}

The exceptions are that for MUNIT and CUT, training for too long resulted in drastic changes in image content for breast MRI and lumbar so we chose earlier model iterations of $10,000$ and $20,000$, respectively for MUNIT, and $20$ epochs for both for CUT.

\subsection{Downstream Task Models}
\label{app:training:downstream}

In this section we describe the architectural and training details of all models trained for the downstream tasks of each dataset (Table \ref{tab:datasets}) on its respective target domain data from the training set. All models are trained with Adam \citep{adam} and a weight decay strength of $10^{-4}$ for $100$ epochs.

\paragraph{Segmentation}
For all segmentation downstream tasks we train a standard UNet \citep{unet} with five encoding blocks, at a batch size of $8$ with a learning rate of $0.01$. The model is trained with equally-weighted cross-entropy and Dice losses, the latter implemented with MONAI \citep{Cardoso_MONAI_An_open-source_2022}.

\paragraph{Object Detection}
For detection downstream tasks, we trained a Faster-RCNN \citep{girshick2015fast} with a batch size of 4 and a learning rate of $0.005$. The model is implemented using Torchvision \citep{torchvision2016}, with the number of predicted classes modified to 2. The loss function is the default loss from this built-in model.

\paragraph{Classification/Slice-level Detection}
For classification tasks we train a standard ResNet-18 \citep{he2016resnet}, modified to take in one-channel inputs and output one logit (as all tasks are binary classification). We use a batch size of $64$ and a learning rate of $0.001$, with a cross-entropy loss.

\section{Additional Experiments}
\subsection{Ablation Studies}
\subsubsection{Radiomic Feature Importance}
\label{sec:app:featureimportance}
\begin{figure}[thpb]
    \centering
    \includegraphics[width=0.99\linewidth]{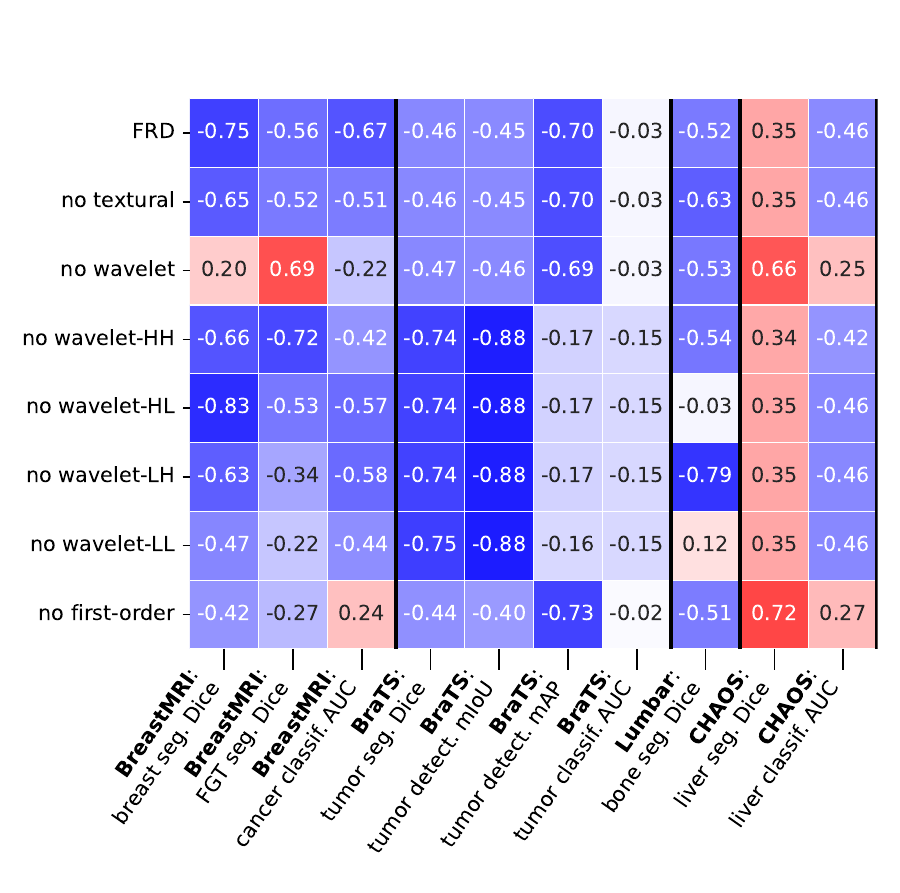}
    \caption{\textbf{Importance of different radiomic features for FRD.} Pearson correlation $r$ between FRD and downstream task performance metrics across all translation models (as in Fig. \ref{fig:metric_corr}), comparing using standard FRD with all features (top row) to removing certain groups of features (lower rows).}
    \label{fig:metric_corr_exclusions}
\end{figure}

To better interpret FRD, we assess the importance of different radiomic feature groups (textural/gray-level matrix, wavelet, first-order) by ablation: examining how removing each group affects the translation model downstream task performance results (Fig. \ref{fig:metric_corr}). For the wavelet features, which correspond to those where the image is passed through a frequency-based filter (-LL, -LH, -HL or -HH) before the computation of the feature, we evaluate both completely excluding all features (``no wavelet'') as well as removing those for only one filter type (-LL, -LH, -HL or -HH).  Results are shown in Fig. \ref{fig:metric_corr_exclusions}.

Overall, wavelet and first-order features are most crucial for FRD, as excluding them significantly worsens correlation results. In evaluating sub-types of wavelet features, we see that excluding wavelet-HH improves performance for three tasks (breast MRI segmentation (FGT) and brain MRI/BraTS segmentation and detection mIoU), yet worsens performance for others (breast MRI segmentation (breast) and classification, BraTS detection mAP, and CHAOS classification). We see similar tradeoffs for excluding wavelet-HL, -LH, or -LL features. In general, there is no consistent advantage to excluding one type of wavelet feature while including the others.

Textural features are somewhat important for breast MRI but have limited impact on other datasets. Breast MRI is generally the most sensitive to feature exclusion, suggesting that subtle domain shifts require a broader range of features for accurate analysis. Overall, these findings indicate that including all types of radiomic features for computing FRD results in a better general-purpose metric.

\subsubsection{Using MMD instead of \Frechet{} Distance}
\label{sec:app:mmd}

\begin{figure}[thpb]
    \centering
    \includegraphics[width=0.99\linewidth]{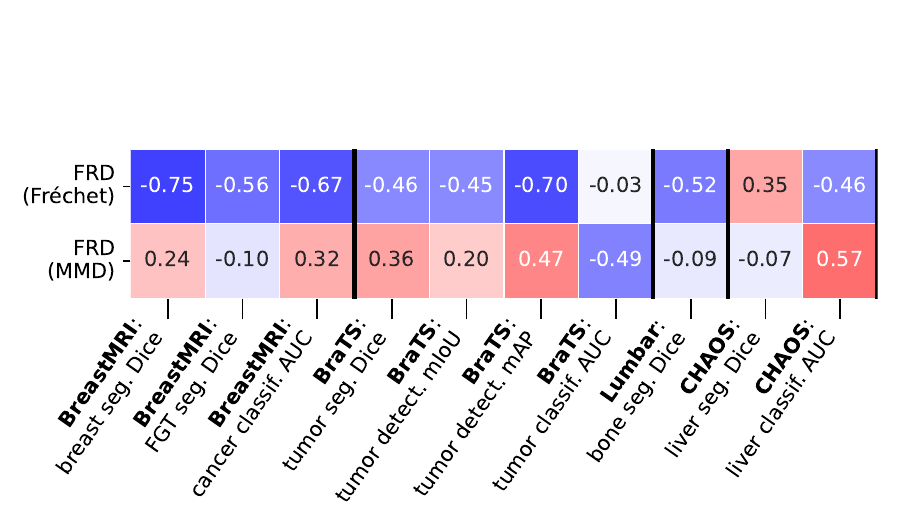}
    \caption{Pearson correlation $r$ between FRD and downstream task performance metrics across all translation models (as in Fig. \ref{fig:metric_corr}), comparing using \Frechet{} or MMD distance for FRD.}
    \label{fig:metric_corr_MMD}
\end{figure}

Perceptual metrics such as CMMD and KID use the MMD (Maximum Mean Discrepancy) distance metric $d_\mathrm{MMD}$ \citep{gretton2012kernel} over the more common \Frechet{} distance, due to advantages such as lacking the Gaussianity assumption and being suitable for smaller datasets \citep{jayasumana2023rethinking,binkowski2018demystifying}. Here we will evaluate calculating our proposed FRD distance using MMD (with a standard Gaussian RBF kernel) as
\begin{equation}
    \label{eq:radiomicmmd}
    \text{FRD}_{\text{MMD}}(D_1, D_2) := d_\mathrm{MMD}(f_{radio}(D_1), f_{radio}(D_2)),
\end{equation}
instead of via \Frechet{} distance as $d_\mathrm{radio}$ (Eq. \eqref{eq:radiomicdist}). We compare the two metrics (FRD and ``FRD-MMD'') in terms of (1) how much they correlate with downstream task performance metrics (as in Fig. \ref{fig:metric_corr}), in Fig. \ref{fig:metric_corr_MMD}, and (2) whether they rank translation models similarly (linearly or non-linearly), in Table \ref{tab:mmd}.

\begin{table}[htbp]
\centering
\begin{tabular}{l|cccc}
\textbf{\tworow{Corr.}{type}} & \textbf{\tworow{Breast}{MRI}} & \textbf{\tworow{Brain}{MRI}} & \textbf{Lumbar} & \textbf{CHAOS} \\
\midrule
Pearson & -0.75 & -0.79 & 0.81 & -0.84 \\
Spearman & -0.83 & -0.83 & 0.49 & -0.54 \\
\bottomrule
\end{tabular}
\caption{Correlation $r$ of FRD computed with MMD distance (Eq. \eqref{eq:radiomicmmd}) with standard \Frechet{} distance FRD (Eq. \eqref{eq:radiomicdist}), across all translation models.}
\label{tab:mmd}
\end{table}

We first see that FRD-MMD is noticeably inferior to FRD in terms of its negative correlation to downstream task performance (Fig. \ref{fig:metric_corr_MMD}); for all but one task, the correlation $r$ is either close to zero, or in the wrong direction (positive $r$, as higher perceptual distance should correlate with worse performance, not better). Moreover, FRD-MMD is not consistent in terms of its relationship to standard FRD (Table \ref{tab:mmd}). We hypothesize that these issues could potentially be due to the dependence of MMD on the choice of kernel, which could require further tuning, or the fact that MMD does \textit{not} have an assumption of Gaussianity unlike the \Frechet{} distance, which may result in a metric that is too unconstrained.

\subsection{OOD Performance Drop Prediction}
\label{app:results_tasksOODdrop}

In Table \ref{tab:results_tasksOODdrop} we evaluate if images detected as out-of-domain with our OOD score thresholding approach (Sec. \ref{sec:ood}) also result in lowered performance compared to on detected ID cases.

\begin{table*}[htbp]
    \centering
    \fontsize{8pt}{8pt}\selectfont
    \begin{tabular}{l|cc|c||c|cc|c||c||cc}
    \multicolumn{1}{c}{} & \multicolumn{3}{c||}{\textbf{Breast MRI}} & \multicolumn{4}{c||}{\textbf{Brain MRI}} &  \multicolumn{1}{c||}{\textbf{Lumbar}} & \multicolumn{2}{c}{\textbf{CHAOS}}  \\
    \toprule
    \multicolumn{1}{c|}{} & \multicolumn{2}{c|}{Dice} &  AUC& \multicolumn{1}{c|}{Dice} & mIoU& mAP & AUC& \multicolumn{1}{c||}{Dice} & \multicolumn{1}{c}{Dice} & AUC\\
    \midrule 
    \textbf{Detected as:} & Breast & FGT & \multicolumn{1}{c|}{Cancer} & Tumor & \multicolumn{2}{c|}{Tumor}  & Cancer & Bone & Liver & Liver  \\
    \midrule
    In-Domain & 0.904 & 0.698 & 0.670 & 0.434 & 0.179 & 0.175 & 0.619 & 0.856 & 0.848 & 0.789 \\
    Out-of-Domain & 0.731 & 0.473 & 0.535 & 0.498 & 0.215 & 0.223 & 0.618 & 0.001 & 0.129 & 0.463 \\
    \bottomrule
    \end{tabular}
    
    \caption{Downstream task performance on test points detected as ID vs. OOD using our thresholding method.}
    \label{tab:results_tasksOODdrop}
\end{table*}

\subsection{OOD Performance Drop Severity Ranking}
\label{app:ood_drop_full}
Here, we assess how well FRD and other perceptual metrics predict performance drops on out-of-domain (OOD) data. Given a model trained on target domain data \( \Dt \) and two new OOD datasets \( D_{\mathrm{OOD},1} \) and \( D_{\mathrm{OOD},2} \), we examine if a metric \( d \) can correctly indicate which OOD dataset will suffer a greater performance drop. Specifically, we test if \( \mathrm{Perf}(D_{\mathrm{OOD},2}^\mathrm{test}) < \mathrm{Perf}(D_{\mathrm{OOD},1}^\mathrm{test}) \) aligns with \( d(D_{\mathrm{OOD},2}^\mathrm{test}, \Dt) > d(D_{\mathrm{OOD},1}^\mathrm{test}, \Dt) \), and vice versa.

We evaluate this scenario with the datasets in Table \ref{tab:datasets} which possess additional data domains beyond the target domain and default source domain $\Ds$, namely, BraTS using its T2-FLAIR data \citep{brats}, and CHAOS using its T1 Dual Out-Phase and T2 SPIR MRI data \citep{kavur2021chaos}. We show these task performance vs. perceptual distance agreement results in Table \ref{tab:ood_perf} for each type of downstream task, and for each possible pair of $D_{\mathrm{OOD},1}$ and $D_{\mathrm{OOD},2}$ for each dataset (T1 MRI and T2 FLAIR MRI for BraTS, respectively, and all $2$-combinations of $\{$T1 Dual Out-Phase MRI, T2 SPIR MRI, and CT$\}$ for CHAOS). Shown in Tables \ref{tab:ood_perf_full_brats} and \ref{tab:ood_perf_full_chaos} are the specific results that generated Table \ref{tab:ood_perf}.

\begin{table*}[htbp]
\centering
\scriptsize


\begin{tabular}{l||c|ccc}
\multicolumn{5}{c}{\textbf{Segmentation} (\textit{Dice})} \\
& \multicolumn{1}{c|}{\tworow{Brain MRI:}{trained on T2}} & \multicolumn{3}{c}{CHAOS: trained on T1 Dual In-Phase} \\
\toprule
 & \tworow{T1}{vs. T2 FLAIR} & \tworow{T1 DOP}{vs. T2 SPIR} & \tworow{T1 DOP}{vs. CT} & \tworow{T2 SPIR}{vs. CT} \\
\midrule
\textbf{FRD} & \cmark & \xmark & \cmark & \cmark \\
\textbf{RadFID} & \cmark & \cmark & \cmark & \cmark \\
\textbf{FID} & \xmark & \cmark & \cmark & \cmark \\
\textbf{KID} & - & \cmark & \cmark & \cmark \\
\textbf{CMMD} & \xmark & \cmark & \cmark & \cmark \\
\bottomrule
\end{tabular}
\hspace{0.4cm}
\begin{tabular}{cc}
\multicolumn{2}{c}{\textbf{Detection}} \\
\multicolumn{2}{c}{\tworow{Brain MRI:}{trained on T2}}\\
\toprule
\multicolumn{2}{c}{T1 vs. T2 FLAIR} \\
\textit{mIoU} & \textit{mAP} \\
\midrule
\xmark & \cmark \\
\xmark & \cmark \\
\cmark & \xmark \\
- & - \\
\cmark & \xmark \\
\bottomrule
\end{tabular}
\hspace{0.4cm}
\begin{tabular}{c|ccc}
\multicolumn{4}{c}{\textbf{Classification} (\textit{AUC})} \\
\multicolumn{1}{c|}{\tworow{Brain MRI:}{trained on T2}} & \multicolumn{3}{c}{CHAOS: trained on T1 Dual In-Phase} \\
\toprule
\tworow{T1}{vs. T2 FLAIR} & \tworow{T1 DOP}{vs. T2 SPIR} & \tworow{T1 DOP}{vs. CT} & \tworow{T2 SPIR}{vs. CT} \\
\midrule
\cmark & \cmark & \cmark & \cmark \\
\cmark & \xmark & \cmark & \cmark \\
\xmark & \xmark & \cmark & \cmark \\
- & \xmark & \cmark & \cmark \\
\xmark & \xmark & \cmark & \cmark \\
\bottomrule
\end{tabular}

\caption{\textbf{Can FRD predict OOD performance drop severity?} For each downstream task type (sub-tables) trained on a given dataset's target domain data $\Dt$ (first row) and for two OOD test sets $D^\mathrm{test}_{\mathrm{OOD},1}$ and $D^\mathrm{test}_{\mathrm{OOD},2}$ (second row), whether $\mathrm{Perf}(D^\mathrm{test}_{\mathrm{OOD},2})<\mathrm{Perf}(D^\mathrm{test}_{\mathrm{OOD},1})$ does (\cmark) or does not (\xmark) correspond to $d(D^\mathrm{test}_{\mathrm{OOD},2}, \Dt)>d(D^\mathrm{test}_{\mathrm{OOD},1}, \Dt)$ and vice-versa. ``-'' denotes that the given perceptual metric was only negligibly affected by the change of the OOD dataset. ``DOP'' is ``Dual Out-Phase''}
\label{tab:ood_perf}
\end{table*}

\begin{table*}[htbp!]
\centering
\small
\begin{tabular}{l|cccc|ccccc}
\multicolumn{1}{c}{} & \multicolumn{4}{c}{\textbf{Downstream task performance}} & \multicolumn{5}{c}{\textbf{Perceptual distance metrics}} \\
$D^\mathrm{test}_\mathrm{OOD}$ & Dice   & mIoU   & mAP    & AUC    & FRD  & RadFID & FID    & KID   & CMMD  \\
\midrule
T1 MRI   & 0.005  & 0.152  & 0.065  & 0.727  & 6.18 & 0.25   & 108 & 0.089 & 0.179 \\
T2 FLAIR MRI & 0.286 & 0.144 & 0.108 & 0.885 & 5.09 & 0.19   & 117 & 0.088 & 0.394 \\
\bottomrule
\end{tabular}
\caption{Downstream task performance $\mathrm{Perf}(D^\mathrm{test}_\mathrm{OOD})$ (left block) and perceptual distances $d(D^\mathrm{test}_\mathrm{OOD}, \Dt)$ (right block) on out-of-domain data $D^\mathrm{test}_\mathrm{OOD}$ (each row) for downstream task models trained on (in-domain) BraTS T2 MRI data $\Dt$, to supplement Table \ref{tab:ood_perf}.}
\label{tab:ood_perf_full_brats}
\end{table*}

\begin{table*}[htbp!]
\centering
\small
\begin{tabular}{l|cc|ccccc}
\multicolumn{1}{c}{} & \multicolumn{2}{c}{\textbf{Downstream task performance}} & \multicolumn{5}{c}{\textbf{Perceptual distance metrics}} \\
$D^\mathrm{test}_\mathrm{OOD}$ & Dice & AUC    & FRD  & RadFID & FID    & KID   & CMMD  \\
\midrule
T1 Dual Out-Phase MRI & 0.779 & 0.853 & 7.87 & 0.09 & 143 & 0.096 & 0.205 \\
T2 SPIR MRI           & 0.262 & 0.867 & 7.55 & 0.20 & 189 & 0.126 & 0.507 \\
CT               & 0.062  & 0.504  & 60.6 & 0.65 & 277 & 0.268 & 1.666 \\
\bottomrule
\end{tabular}
\caption{Downstream task performance $\mathrm{Perf}(D^\mathrm{test}_\mathrm{OOD})$ (left block) and perceptual distances $d(D^\mathrm{test}_\mathrm{OOD}, \Dt)$ (right block) on out-of-domain data $D^\mathrm{test}_\mathrm{OOD}$ (each row) for downstream task models trained on (in-domain) CHAOS T1 Dual In-Phase MRI data $\Dt$, to supplement Table \ref{tab:ood_perf}.}
\label{tab:ood_perf_full_chaos}
\end{table*}

\subsection{Towards Dataset-Level OOD Detection}
\label{app:ood_datasetlevel}
In Sec. \ref{sec:ood}, we showed how FRD/radiomic features can be used for single image-level binary OOD detection. However, a more realistic scenario may be that some new dataset is acquired from an outside hospital/site, and we wish to know if the dataset is generally OOD relative to our own reference ID dataset $D_\mathrm{ID}$ that we used to train some downstream task model, to get some idea of how our model will perform on the new dataset $D_\mathrm{test}$. For example, our ID dataset could be breast MRI collected from GE scanners, and the new dataset could potentially have OOD (\eg, Siemens) images. Our goal is therefore to have a metric that returns an (approximately) standardized value if $D_\mathrm{test}$ is OOD. 

A naive prior approach to this could be to measure the FID or RadFID between $D_\mathrm{ID}$ and $D_\mathrm{test}$, but as we will show, such distances are not clearly interpretable due to the distance value being noticeably affected by the specific dataset used, as well as the sample size (Sec. \ref{sec:sampleeff}). To this end, we propose a FRD-based metric for dataset-level OOD detection which is designed to return $1$ (or a value close to it) when the test set is completely OOD. 

We do so by considering an ID reference point $x_\mathrm{ID}\sim D_\mathrm{ID}$ and test set point $x\sim D_\mathrm{test}$, both randomly sampled. Now, we wish to have a metric that estimates the probability that the test set is OOD. The key insight here is that the higher this probability, the higher the chance that $x$ is OOD, such that it's expected score/distance from $D_\mathrm{ID}$, $s(x)$ (Eq. \ref{eq:ood}) will in turn be more likely to be larger than that of a typical ID point $x_\mathrm{ID}$. Assuming that OOD points will not be typically \textit{closer} to $D$ than ID points, which is true by the definition of OOD, then the minimum value of this probability is $\mathrm{Pr}[s(x) > s(x_\mathrm{ID})] = 0.5$ if $D_\mathrm{test}$ is 100\% ID (no clear difference between the test set and reference set score distributions), and $\mathrm{Pr}[s(x) > s(x_\mathrm{ID})] = 1$ if $D_\mathrm{test}$ is 100\% OOD.

We then convert this to a metric, $\text{nFRD}_\mathrm{group}$ (FRD for group-level OOD detection normalized to a fixed range) that ranges from $0$ to $1$ with
\begin{equation}
    \text{nFRD}_{\mathrm{group}} := 2(\mathrm{Pr}[s(x) > s(x_\mathrm{ID})] - 0.5).
\end{equation}
The final question is then how $\mathrm{Pr}[s(x) > s(x_\mathrm{ID})]$ can be computed in practice; thankfully, the area under the ROC curve (AUC) \textit{by definition} is this quantity \citep{fawcett2006introduction}, which can be easily computed, giving
\begin{equation}
    \label{eq:nradgroup}
    \text{nFRD}_{\mathrm{group}} := 2(\mathrm{AUC}[S_\mathrm{test}, S_\mathrm{ID}] - 0.5),
\end{equation}
where $S_\mathrm{test} := \{ s(x) : x \in D_\mathrm{test}\}$ and $S_\mathrm{ID}$ is the reference distribution of ID scores, $S_\mathrm{ID} := \{s(x_\mathrm{ID};D_\mathrm{ID} \setminus x_\mathrm{ID}) : x_\mathrm{ID} \in D_\mathrm{ID}\}$, as in Sec. \ref{sec:ood}.

We evaluate $\text{nFRD}_\mathrm{group}$ for OOD-scoring OOD test sets in Table \ref{tab:ood_group_onOOD}, averaged over 10 randomly sampled test sets of size $100$ for each trial, compared to using the FID or FRDFID between $D_\mathrm{test}$ and $D_\mathrm{ID}$. While all metrics assign a higher score for the OOD test set than the ID test set, we note that the scale of FID and RadFID OOD test set distance values changes noticeably depending on the dataset, a factor which would be even more pronounced if considering datasets of different sizes, as those metrics can be highly unstable for different sample sizes (Sec. \ref{sec:sampleeff}). On the other hand, $\text{nFRD}_\mathrm{group}$ is $\approx 1$ for the OOD test set in 3/4 datasets (besides BraTS, due to it generally proving difficult for disentangle the ID and the OOD distributions (Sec. \ref{sec:ood})), making it a more standardized, interpretable and practical metric. This enables us to posit that a $\text{nFRD}_\mathrm{group}$ score of $\approx 1$ for some new dataset means that the dataset is likely OOD.

We similarly see that for completely ID test sets (Table \ref{tab:ood_group_onID}), $\text{nFRD}_\mathrm{group}$ is $\approx 0$ in $3/4$ cases. While RadFID does so for $4/4$ cases, this doesn't account for the fact that RadFID is still highly sensitive to sample size, hurting its interpretable, standardized, realistic use in this case.
Finally, we also show ablation studies in these two tables of using ImageNet or RadImageNet features to compute $\text{nFRD}_\mathrm{group}$ instead of radiomic features, where we see that using these learned features results in less stable OOD test set distance values.

\begin{table}[htbp]
\setlength{\tabcolsep}{3pt}
\fontsize{8pt}{8pt}\selectfont
\centering
\begin{tabular}{l|cccc}
\textbf{Metric} & \textbf{\tworow{Breast}{MRI}} & \textbf{\tworow{Brain}{ MRI}} & \textbf{\tworow{}{Lumbar}} & \textbf{\tworow{}{CHAOS}} \\
\midrule
FID & 178   & 223  & 277  & 338  \\
RadFID & 0.22  & 0.35 & 1.23 & 1.64 \\
$\text{nFRD}_{\mathrm{group}}$ & 1.00 & 0.62 & 0.95 & 1.00    \\
$+$ImageNet & 0.01  & 0.84 & 0.75 & 0.94 \\
$+$RadImageNet & 0.14 & 0.32 & 0.99 & 0.94 \\
\bottomrule
\end{tabular}
\caption{Dataset-level OOD detection scores for OOD test sets.}
\label{tab:ood_group_onOOD}
\end{table}

\begin{table}[htbp]
\setlength{\tabcolsep}{3pt}
\fontsize{8pt}{8pt}\selectfont
\centering
\begin{tabular}{l|cccc}
\textbf{Metric} & \textbf{\tworow{Breast}{MRI}} & \textbf{\tworow{Brain}{ MRI}} & \textbf{\tworow{}{Lumbar}} & \textbf{\tworow{}{CHAOS}} \\
\midrule
FID & 92   & 73    & 77   & 48    \\
RadFID & 0.09 & 0.06  & 0.04 & 0.04  \\
$\text{nFRD}_{\mathrm{group}}$ & 0.00    & 0.04  & 0.07 & 0.44  \\
$+$ImageNet & 0.22 & -0.04 & 0.05 & -0.05 \\
$+$RadImageNet & 0.1  & -0.05 & 0.07 & -0.13 \\
\bottomrule
\end{tabular}
\caption{Dataset-level OOD detection scores for ID test sets.}
\label{tab:ood_group_onID}
\end{table}

\subsection{FRD vs. FRD$_\text{\textbf{v0}}$: a Study on Realistic MRI Corruptions}
\label{app:MRIdistort}
Now we will compare the sensitivity of FRD to image corruptions compared to its predecessor, FRD$_\text{\textbf{v0}}$ \cite{frd}, in the context of realistic MRI corruptions. Given an image dataset $D$, we will measure the effect that randomized transformations $T$ have on the distance $d(D, T(D))$ (measured via FRD or FRD$_\text{\textbf{v0}}$), as $T$ becomes more severe. The dataset that we use is the CHAOS T1-Dual In-Phase MRI training set (Table. \ref{tab:datasets}). These transformations (besides random swapping) were chosen to simulate realistic artefacts to which MRI may be susceptible, due to issues such as motion, noise, etc.

Given some image $x\sim D$ with pre-established maximum intensity $I_{max}$, the transformations $T$ that we consider are listed as follows, each controlled by a percentage/severity parameter $p\in[0,100]$.
\begin{enumerate}
    \item \textbf{Gaussian noise.} Random Gaussian noise is added to $x$ via $x\leftarrow \mathrm{clip}(x + \frac{pI_{max}}{100}\epsilon, [0, I_{max}])$ where $\epsilon\sim \mathcal{N}(0, 1)$.
    \item \textbf{Gaussian blur.} Apply Gaussian blur with kernel size $k=\frac{p}{100}\times\texttt{max\_size}$, rounded to the nearest odd integer, where $\texttt{max\_size}$ is the pixel length of the longest side of the image.
    \item \textbf{Random swap.} Randomly chosen small square patches of size $k=15$ are swapped in the image, repeated $\mathrm{round}(p)$ times.
    \item \textbf{Motion artifacts.} MRI motion artifacts are simulated with realistic movement transformations (rotation and/or translation) via the \texttt{RandomMotion} function of TorchIO \citep{perez-garcia_torchio_2021}. We used \texttt{RandomMotion} with parameters $\texttt{degrees} = \frac{p}{100}\times 10$, $\texttt{translation} =\frac{p}{100}\times 10$, and $\texttt{num\_transforms} = \mathrm{max}(1, \mathrm{round}(\frac{p}{100}\times 2))$.
    \item \textbf{Random bias field.} MRI bias field artifacts are simulated via spatially-varying low-frequency intensity variations, implemented via TorchIO's \texttt{RandomBiasField} function with $\texttt{coefficients}=\frac{p}{100}\times0.5$.
    \item \textbf{Deformation.} MRI soft tissue deformations are simulated via random non-linear spatial corruptions, using TorchIO's \texttt{RandomElasticDeformation} function. We use parameters of $\texttt{max\_displacement} = \frac{p}{100}\times 7.5$ and $\texttt{num\_control\_points} = \mathrm{max}(5, \mathrm{round}(\frac{p}{100} \times 7))$.
\end{enumerate}

\begin{figure}[thpb]
    \centering
    \includegraphics[width=0.95\linewidth]{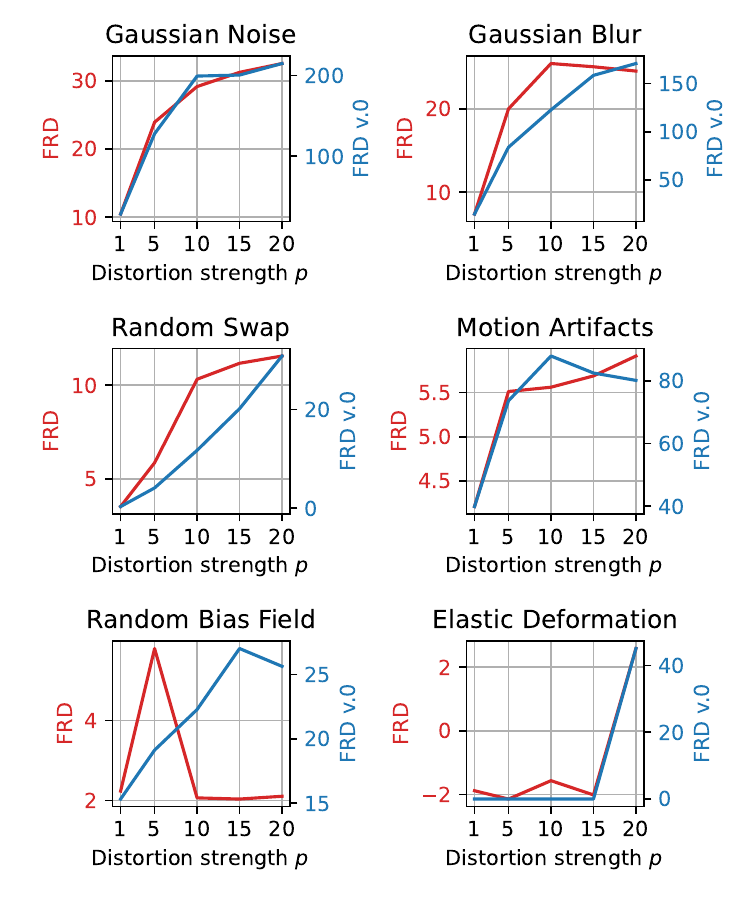}
    \caption{Comparison of FRD and FRD$_\text{\textbf{v0}}$ \citep{frd} in terms of sensitivity to abdominal MR image corruptions.}
    \label{fig:distortions_frdcompare}
\end{figure}

We plot the results of this study in Fig. \ref{fig:distortions_frdcompare}. We see that FRD and FRD$_\text{\textbf{v0}}$ have similar sensitivity behavior to image corruptions, besides the random bias field results for $p=5$. In general, the two distance metrics increase as the corruption severity increases, which is the desired behavior.

\subsection{Adversarial Attack Success Results}
\label{app:egattacks}
\begin{figure}[thpb]
    \centering
    \includegraphics[width=0.85\linewidth]{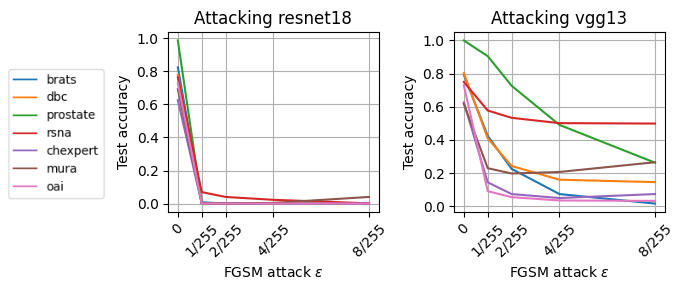}
    \includegraphics[width=0.95\linewidth]{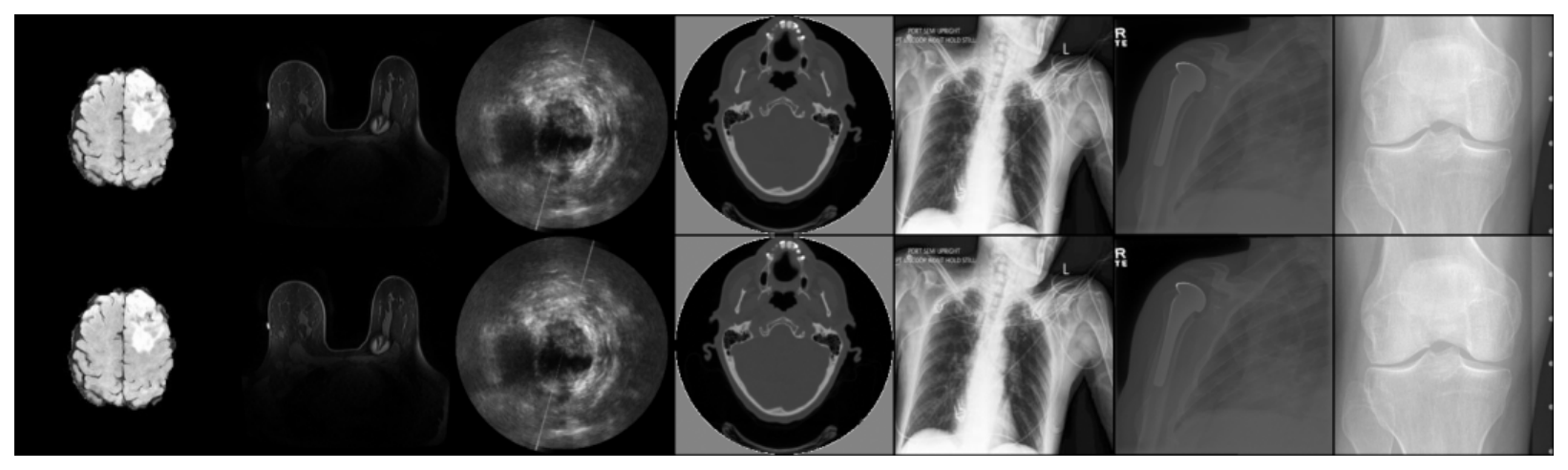}
    \caption{\textbf{Top:} Adversarial attack success (accuracy drop) on ResNet-18 and VGG-13 classification models. \textbf{Bottom:} Example non-attacked (first row) and attacked (second row) versions of the same images, from $\epsilon=1/255$ FGSM attacks on ResNet-18 models for different datasets (columns of image; ordered left-to-right by legend labels in the top plots).}
    \label{fig:advatk_success}
\end{figure}

We show the success of our adversarial attacks in terms of decreased model accuracy in Fig. \ref{fig:advatk_success}, alongside example clean and attacked images.

\subsection{Calibrating User Preference Ratings}
\label{app:readerstudycalibrated}
\begin{figure}
\centering
\begin{tabular}{l||ccc}
    \multicolumn{1}{c}{} & \multicolumn{3}{c}{\textbf{Absolute Rating}} \\
    \textbf{Metric} & Pearson $r$ & Spearman $r$ & Kendall $\tau$ \\\toprule
    FRD++ & -0.44 & -0.49 & -0.33 \\
    FRD$_\text{\textbf{v0}}$ & -0.66 & -0.6 & -0.47 \\
    FID & 0.32 & 0.6 & 0.47 \\
    RadFID & 0.67 & 0.66 & 0.6 \\
    \bottomrule
\end{tabular}
\includegraphics[width=0.95\linewidth]{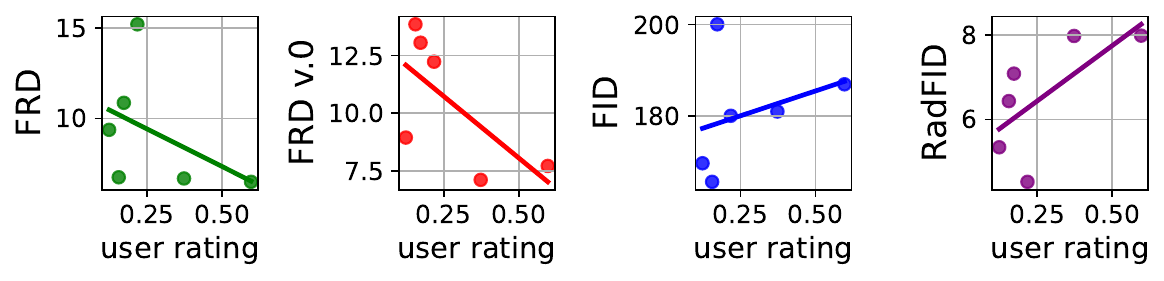}
\caption{\textbf{Top:} Correlation coefficients (linear Pearson $r_P$ and non-linear/rank Spearman $r_S$ and Kendall $\tau$) of different distance metrics with \textbf{calibrated} average user (radiologist) preference, for the task of measuring synthetic image quality.  \textbf{Bottom:} Associated plots and linear best fits for this data.}
\label{fig:userstudy_calib}
\end{figure}

Continuing from Sec. \ref{sec:userstudy}, a related way to measure a reader's rating of the quality of synthetic images is not just by their absolute average Likert score for the images, but by calibrating this measure by subtracting the average score for a set of relevant real images from it (in particular, 15 random real images for a given dataset and anatomical view). This aims to mitigate confounding effects on user preference, such as certain readers having a general bias in rating both synthetic and real images, higher or lower. We show the correlation results of this in Fig. \ref{fig:userstudy_calib}.

Similar to Fig. \ref{fig:userstudy}, we see that FRD correlates (negatively) fairly well with calibrated user preference, while FID and RadFID actually, undesirably, correlate positively.
While FRD is substantially more correlated with user preference than FID and RadFID, FRD$_\text{\textbf{v0}}$ does show a somewhat stronger correlation compared to FRD. 
Intuitively, this is plausible due to FRD including numerous additional frequency/wavelet radiomic features that are not present in FRD$_\text{\textbf{v0}}$, many of which (those not at low frequencies) are typically barely perceptible by the human eye \citep{dieleman2020typicality,salimans2017pixelcnn}; thus, FRD will focus more on such features when comparing images. 
As demonstrated, the inclusion of these features resulted in noticeable improvements to FRD over FRD$_\text{\textbf{v0}}$ in many medical image analysis applications (see Sec. \ref{sec:experiments} and \ref{sec:app:featureimportance}), \ie, where accounting for these subtle details is helpful for the underlying task. 

\subsection{Attempting to Interpret Differences between Medical Image Distributions with Learned Features}
\label{app:interp_featureinversion}
We have applied feature inversion \citep{olah2017feature, featureinvert} to visualize \( \Delta h \) using ImageNet and RadImageNet features (via Lucent \citep{noauthor_greentfrapplucent_nodate}), for breast MRI and brain MRI UNSB translation models, shown in Fig. \ref{fig:translation_inversion}. While the results hint at general textural and shape changes from translation, they lack clear, quantitative insights useful for clinical interpretation.

\section{Additional Discussion}
\subsection{Downstream Task Metrics as Image Distribution Distance Metrics}
\label{sec:app:taskmetrics_are_distmetrics}

Segmentation performance metrics are themselves distance functions between two distributions of image features. If the predictions of the downstream task model on its test set and the corresponding ground truth labels/segmentations are taken as ``features'' of the images, then performance metrics such as Dice segmentation coefficient, IoU, etc. are image distribution metrics that clearly follow the topological requirements of distance metrics: reflexivity, non-negativity, symmetry, and the triangle inequality (Sec \ref{sec:relwork}).


\end{document}